\documentclass[letterpaper, 10 pt, conference]{ieeeconf}  % Comment this line out if you need a4paper

\IEEEoverridecommandlockouts                              % This command is only needed if 
\usepackage{geometry}
\usepackage{graphicx,psfrag,amsmath,amssymb,amsfonts,verbatim,url,booktabs,array,float}
\usepackage[short]{optidef}
\usepackage{censor}
\usepackage{bm}
\usepackage{subfloat}
\usepackage{sidecap}
\usepackage[hidelinks]{hyperref}
\usepackage{soul}
\usepackage{algorithm}
\usepackage[noend]{algpseudocode}
\usepackage{textcomp}
\usepackage{stfloats}
\usepackage{balance}
\usepackage{multicol}

\usepackage{subcaption}
\usepackage[font=small,labelfont=bf]{caption}
\def\BibTeX{{\rm B\kern-.05em{\sc i\kern-.025em b}\kern-.08em
    T\kern-.1667em\lower.7ex\hbox{E}\kern-.125emX}}

\usepackage{xcolor}
\definecolor{orange}{RGB}{255,127,0}
\definecolor{cyan}{RGB}{0,255,255}
\definecolor{magenta}{RGB}{255,0,255}
\definecolor{nicegreen}{rgb}{0.1, 0.6, 0.2}

\newcommand{\addition}[1]{\textcolor{black}{#1}}

\renewcommand{\baselinestretch}{1}

\usepackage{pgfplots}
\pgfplotsset{compat=1.16}

\usepackage{tikz}
\usepackage{tikzscale}
\usetikzlibrary{matrix}
\usetikzlibrary[pgfplots.groupplots]
\usetikzlibrary{arrows.meta}
\usetikzlibrary{backgrounds}
\usepackage{pgfplots}
\usepgfplotslibrary{groupplots}

\usepackage{booktabs}

\newcommand\blfootnote[1]{%
  \begingroup
  \renewcommand\thefootnote{}\footnote{#1}%
  \addtocounter{footnote}{-1}%
  \endgroup
}

\newcommand{\ones}{\mathbf 1}

\begin{document}

\title{
    % \vspace{-7mm} 
    \LARGE \bf Single-Level Differentiable Contact Simulation
    % \vspace{-5mm}
}

\author{
    Simon Le Cleac'h, Mac Schwager, Zachary Manchester, Vikas Sindhwani, Pete Florence and Sumeet Singh
    \thanks{This work was supported by Google Research. (Corresponding author: Simon Le Cleac'h.}
    \thanks{Simon Le Cleac’h is with the Department of Mechanical Engineering, Stanford University, Stanford, CA 94305 USA, 
        {\tt\footnotesize simonlc@stanford.edu}.}
    \thanks{Mac Schwager is with the Aeronautics and Astronautics Department, Stanford University, Stanford, CA 94305 USA,
		{\tt\footnotesize schwager@stanford.edu}.}
    \thanks{Zachary Manchester is with The Robotics Institute, Carnegie Mellon University, Pittsburgh, PA 15213 USA,
		{\tt\footnotesize zacm@cmu.edu}.}
    \thanks{Sumeet Singh and Vikas Sindhwani are with Robotics at Google, New York City, NY 10011 USA,
        {\tt\footnotesize ssumeet@google.com}, 
        {\tt\footnotesize sindhwani@google.com}.}
    \thanks{Pete Florence is with Robotics at Google, Mountain View, CA 94043 USA,
        {\tt\footnotesize peteflorence@google.com}.}
}

\twocolumn[{%
\renewcommand\twocolumn[1][]{#1}%
\maketitle
\vspace{-10mm}
\begin{center}
		{
		\vspace{0.0cm}
		\normalsize
}
\centering

% \vspace{-3mm}
\href{https://simon-lc.github.io/papers/silico/assets/banner.gif}{
        \begin{tikzpicture}
            \draw (0, 0) node[inner sep=0] {\includegraphics[width=1.0\linewidth]
            {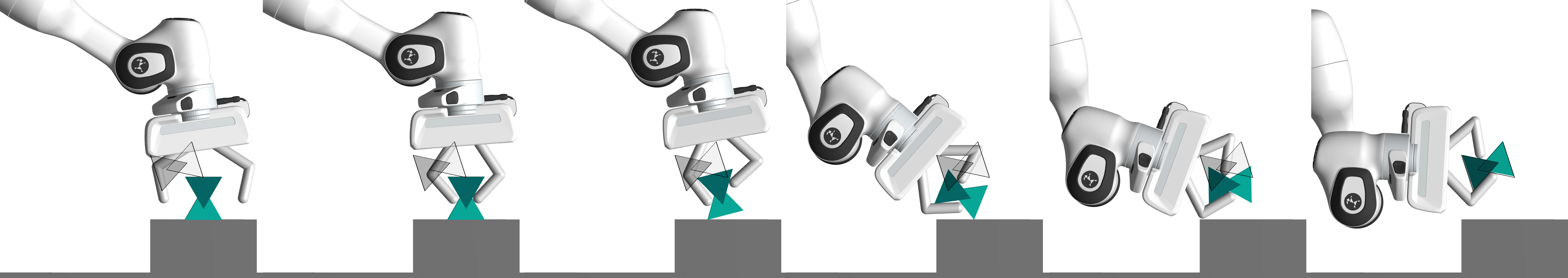}};
            % {figures/test.png}};
            \draw (-6.80, -1.3) node {\textcolor{white}{$\scriptstyle \mathbf{t = 0.0}$\textbf{\scriptsize s}}};
            \draw (-3.70, -1.3) node {\textcolor{white}{$\scriptstyle \mathbf{t = 0.5}$\textbf{\scriptsize s}}};
            \draw (-0.65, -1.3) node {\textcolor{white}{$\scriptstyle \mathbf{t = 1.0}$\textbf{\scriptsize s}}};
            \draw (+2.40, -1.3) node {\textcolor{white}{$\scriptstyle \mathbf{t = 1.5}$\textbf{\scriptsize s}}};
            \draw (+5.45, -1.3) node {\textcolor{white}{$\scriptstyle \mathbf{t = 2.0}$\textbf{\scriptsize s}}};
            \draw (+8.55, -1.3) node {\textcolor{white}{$\scriptstyle \mathbf{t = 2.5}$\textbf{\scriptsize s}}};
        \end{tikzpicture}}
    \vspace{-5mm}
    \captionof{figure}{
    The robot arm is tasked with manipulating the green object to a desired pose (light gray overlay). Our simulation framework provides gradients of both the contact simulation and collision detection, accelerating existing trajectory and policy optimization methods that require gradients. The grasping plan shown is computed in $1$ second on a laptop with an existing planner \cite{pang2022global} using gradients from our simulator. 
    \textbf{Click the image to play the video in a browser.}
    }
    \label{fig:rrt}
\end{center}
}]

\begin{abstract}
    We present a differentiable formulation of rigid-body contact dynamics for objects and robots represented as compositions of convex primitives. Existing optimization-based approaches simulating contact between convex primitives rely on a bilevel formulation that separates collision detection and contact simulation. These approaches are unreliable in realistic contact simulation scenarios because isolating the collision detection problem introduces contact location non-uniqueness. Our approach combines contact simulation and collision detection into a unified single-level optimization problem. This disambiguates the collision detection problem in a physics-informed manner. Compared to previous differentiable simulation approaches, our formulation features improved simulation robustness and a reduction in computational complexity by more than an order of magnitude. We illustrate the contact and collision differentiability on a robotic manipulation task requiring optimization-through-contact. We provide a numerically efficient implementation of our formulation in the Julia language called \href{https://github.com/simon-lc/Silico.jl}{Silico.jl}.
        \blfootnote{This work was supported by Google Research. (Corresponding author: Simon Le Cleac'h.}
        \blfootnote{Simon Le Cleac’h is with the Department of Mechanical Engineering, Stanford University, Stanford, CA 94305 USA, 
            {\tt\footnotesize simonlc@stanford.edu}.}
        \blfootnote{Mac Schwager is with the Aeronautics and Astronautics Department, Stanford University, Stanford, CA 94305 USA,
    		{\tt\footnotesize schwager@stanford.edu}.}
        \blfootnote{Zachary Manchester is with The Robotics Institute, Carnegie Mellon University, Pittsburgh, PA 15213 USA,
    		{\tt\footnotesize zacm@cmu.edu}.}
        \blfootnote{Sumeet Singh and Vikas Sindhwani are with Robotics at Google, New York City, NY 10011 USA,
            {\tt\footnotesize ssumeet@google.com}, 
            {\tt\footnotesize sindhwani@google.com}.}
        \blfootnote{Pete Florence is with Robotics at Google, Mountain View, CA 94043 USA,
            {\tt\footnotesize peteflorence@google.com}.}
\end{abstract}

\section{Introduction}
    Physics engines are increasingly playing a key role in planning and control for robotic manipulation and locomotion tasks. They enable large scale collection of simulated data for reinforcement learning and other policy optimization methods, which can transfer to actual robot hardware~\cite{akkaya2019solving,lee2020learning, rudin2022learning}. They also serve as a testing and validation tool to drastically speed up mechanical design and the deployment on hardware of model-based algorithms such as model predictive control (MPC)~\cite{kuindersma2016optimization}. These acceleration benefits stem from using autodifferentiated gradients, replacing computationally intensive deterministic and stochastic gradient sampling schemes. However, many differentiable physics engines provide subgradient information which is locally exact but myopic about the broader dynamics landscape~\cite{heiden2020neuralsim, lee2018dart, drake} resulting in poor optimization performance. 
    
    More recently Dojo~\cite{howelllecleach2022} introduced an approach to simulate stiff contact while providing smooth analytical gradients. For simple systems, this smooth gradient recovers the gradient obtained by a randomized smoothing sampling scheme in the limit of an infinite number of samples~\cite{pang2022global}. It has proven useful in online trajectory tracking~\cite{lecleach2021fast} for robotic locomotion and global planning~\cite{pang2022global} for contact-rich manipulation. Dojo relies on smooth analytical collision detection, and it has been applied to simple collision geometries: spheres and planes. Our approach extends the capabilities of the implementation provided in \cite{howelllecleach2022} to more complex collision geometries that do not admit analytical collision expressions. 
    
    In this work, we propose a novel formulation of the contact physics that combines optimization-based dynamics \cite{manchester2020variational, howelllecleach2022}
    and optimization-based collision detection~\cite{tracy2022differentiable}. This formulation handles a diverse set of shape primitives such as polytopes, cones, capsules, cylinders, and ellipsoids, thus enabling more accurate geometric modeling of robots and their environments (Fig.~\ref{fig:rrt}). Subsequently, this enables richer contact interaction between the robot and its environment. For instance, a quadruped could exploit contact interaction beyond its four feet by using its torso to hold a door open. A robotic arm could use its links in addition to its end-effector to manipulate large objects.

    % \begin{figure}[t]
    %     \begin{center}
    %         \begin{tikzpicture}
    %             \draw (0, 0) node[inner sep=0] {\includegraphics[width=1.0\columnwidth]{figures/rrt_grasping.png}};
    %             \draw (-2.20, -1.33) node {\textcolor{white}{$\mathbf{t = 0}$\textbf{s}}};
    %             \draw (+0.68, -1.33) node {\textcolor{white}{$\mathbf{t = 1}$\textbf{s}}};
    %             \draw (+3.75, -1.33) node {\textcolor{white}{$\mathbf{t = 2}$\textbf{s}}};
    %         \end{tikzpicture}
    %     \end{center}
    %     \caption{The robot arm is tasked with manipulating the green object to a desired pose (light green). Our simulation framework provides gradients of both the contact simulation and collision detection, accelerating existing trajectory and policy optimization methods that require gradients. The grasping plan shown is computed in $1$ second on a laptop with an existing planner \cite{pang2022global} using gradients from our simulator.}
    %     \label{fig:rrt}
    % \end{figure}

Our key contributions include:
\begin{itemize}
    \item Streamlined sensitivity analysis of optimization-based collision detection 
    \item A novel single-level formulation removing contact point ambiguity and enabling differentiable and reliable simulation of contact dynamics for a variety of shape primitives
    \item An efficient and flexible implementation of the proposed algorithm
    \item A comparison with existing approaches exhibiting a $30\times$ improvement in terms of solve time, and a failure rate decrease from more than $50\%$ down to $0.1\%$.
\end{itemize}

    In the remainder of this paper, we provide an overview of related work on differentiable contact simulation and collision detection in Section~\ref{sec:related_work}. In Section~\ref{sec:technical_background}, we outline important technical background. In Sections~\ref{sec:bilevel_optimization}, \ref{sec:silico}, we introduce the bilevel and our single-level formulation of contact simulation and collision detection. In Section~\ref{sec:experiments}, we implement our approach in a variety of scenarios and compare it to an existing baseline in terms of accuracy, robustness and computational complexity. Finally, in Section~\ref{sec:limitations}, we discuss limitations of our approach and provide closing remarks in Section~\ref{sec:conclusion}.

    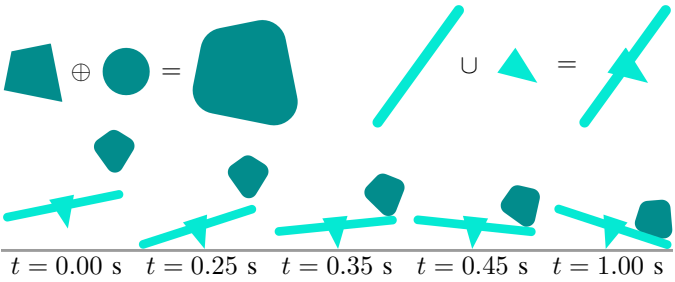
\begin{figure}[t]
        \begin{center}
            \begin{tikzpicture}
                \draw (0, 0) node[inner sep=0] {\includegraphics[width=0.45\columnwidth]{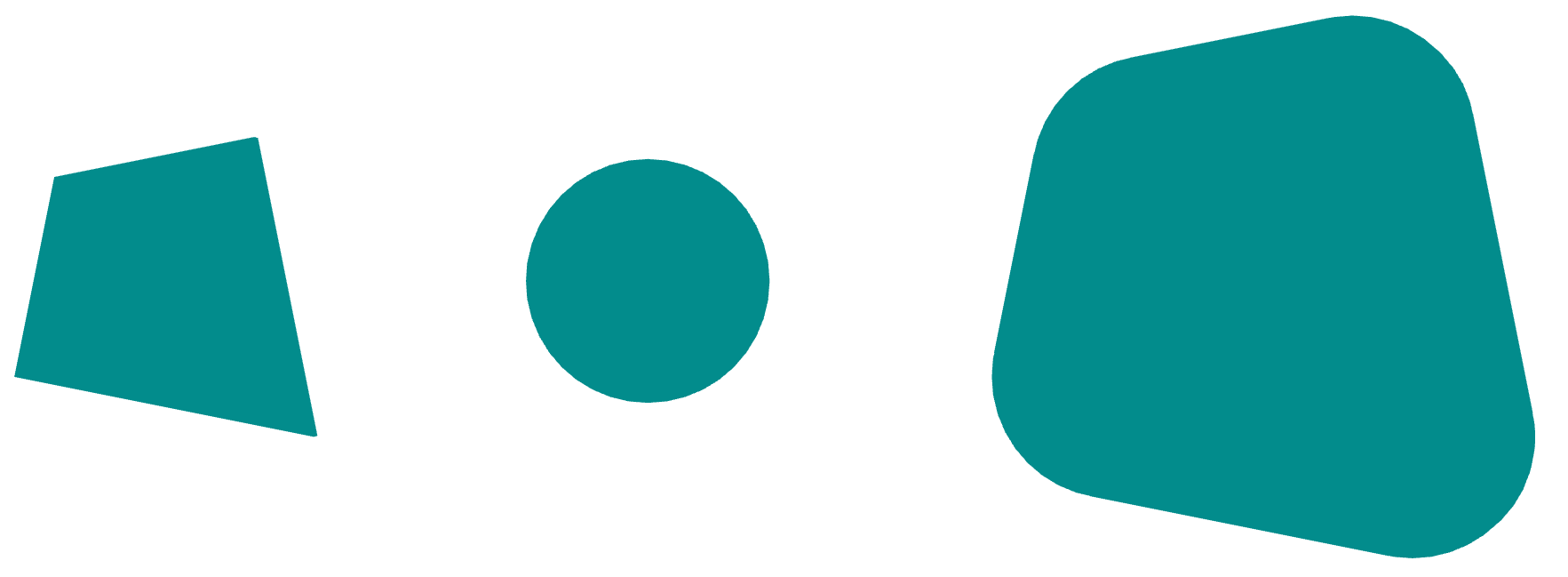}};
                \draw (-0.95, -0.0) node {$\oplus$};
                \draw (+0.25, -0.0) node {$=$};
            \end{tikzpicture}
            \hfill
            \begin{tikzpicture}
                \draw (0, 0) node[inner sep=0] {\includegraphics[width=0.45\columnwidth]{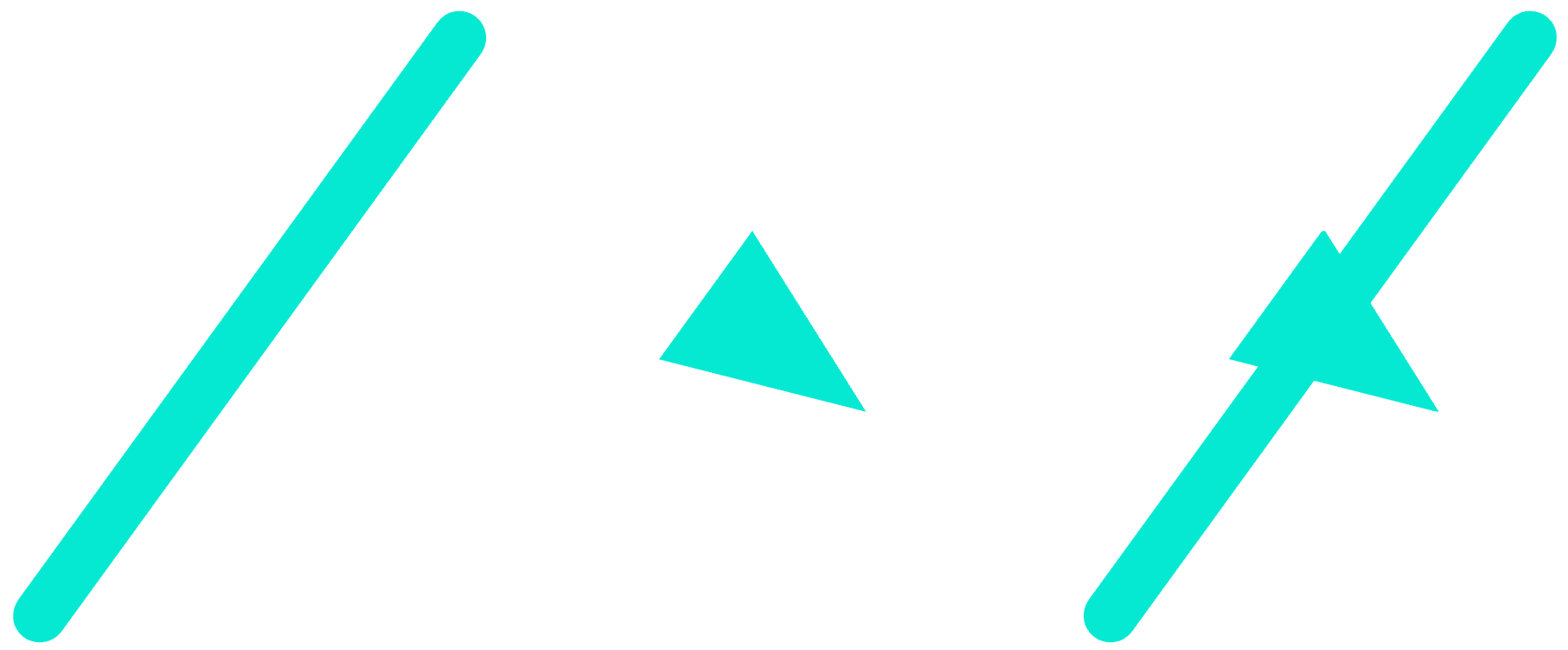}};
                \draw (-0.70, -0.0) node {$\cup$};
                \draw (+0.60, -0.0) node {$=$};
            \end{tikzpicture}
        \end{center}
        \vspace{-4mm}
        \begin{center}
            \begin{tikzpicture}
                \draw (0, 0) node[inner sep=0] {\includegraphics[width=1.0\columnwidth]{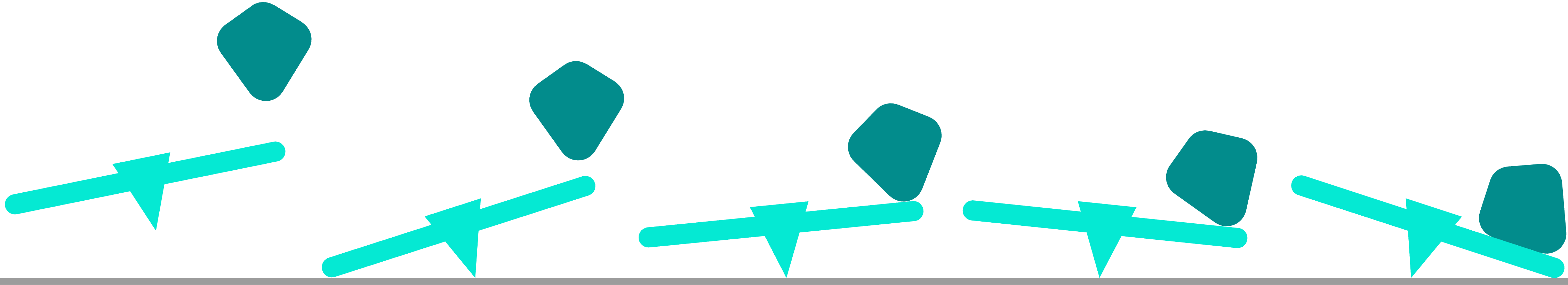}};
                \draw (-3.60, -1.0) node {$t = 0.00$ s};
                \draw (-1.80, -1.0) node {$t = 0.25$ s};
                \draw (+0.00, -1.0) node {$t = 0.35$ s};
                \draw (+1.80, -1.0) node {$t = 0.45$ s};
                \draw (+3.60, -1.0) node {$t = 1.00$ s};
            \end{tikzpicture}
        \end{center}
        \caption{
        We compose simple convex primitives to form non-convex shapes, termed \emph{convex bundles}, through Minowski sums (top left) and union operations (top right). We simulate contact interactions between the two convex bundles and the floor (bottom).}
        \label{fig:convex_bundles}
    \end{figure}

\section{Related Work}
\label{sec:related_work}
    Well established physics engines such as Bullet~\cite{coumans2021}, Drake~\cite{drake}, Mujoco~\cite{todorov2012mujoco}, and DART~\cite{lee2018dart} typically rely on non-differentiable algorithms to perform collision detection. For instance, the Gilbert, Johnson, and Keerthi (GJK) algorithm~\cite{gilbert1988fast} computes the minimum distance between two convex sets usually represented by meshes. Similarly, the Expanding Polytope Algorithm (EPA)~\cite{van2001proximity} computes the penetration depth and a contact normal for overlapping shapes. Because of these non-differentiable modules, these simulators cannot provide analytical gradients and hence need to resort to sampling-based approximations.
    
    Recently, the Tiny Differentiable simulator (TinyDiff) which relies on automatic differentiation was proposed~\cite{heiden2021neuralsim}. However, collision detection is only supported for a reduced set of shape primitives for which contact locations and separating distances can be computed with analytical formulas (e.g. plane-box, plane-sphere, sphere-box). Because the analytic formula contain control flow expressions, they cannot provide smooth gradient information. In this work, we follow, Dojo's~\cite{howelllecleach2022} approach 
    treating simulation as an optimization problem. This provides smooth gradients for contact pairs that are expressed with smooth analytical formulas (e.g. plane-sphere, sphere-sphere).
    
    A randomized-smoothing approach to the collision detection problem~\cite{montaut2022differentiable} computes contact gradients via sampling between pairs of convex meshes. It resorts to classical collision detection algorithms: GJK when the objects are not in collision and EPA otherwise. However, the approach has not been demonstrated in the context of contact simulation. 
    Finally, an optimization-based approach DCOL~\cite{tracy2022differentiable} performs collision detection between a diverse set of shape primitives (polytopes, cones, capsules, etc). 
    This approach has been successfully embedded in a trajectory optimization solver on collision avoidance problems. It has similarly been applied to contact simulation via bilevel optimization where the \emph{upper-level} problem encodes contact dynamics and the \emph{lower-level} problem encodes collision detection. However, the bilevel formulation struggles in situations that are ubiquitous in practical robotics scenarios, specifically when contact between two shapes cannot be reduced to a single point, e.g. a box resting on another box. 
    Our approach instead formulates physics simulation as a single-level optimization problem, thereby unifying the collision detection and contact simulation problems that have historically been treated separately. Importantly, our simulation results demonstrate that this approach drastically improves the reliability of contact simulation as compared to the bilevel formulation \cite{tracy2022differentiable, wang2022linear}.

\section{Technical Background}
\label{sec:technical_background}
In this section, we provide a concise overview of differentiable optimization, followed by background on optimization-based contact simulation and bilevel/single-level optimization.

    \subsection{Differentiable Optimization}
    \label{sec:sensitivity_background}
    
    We consider a constrained optimization problem, 
    \begin{equation}
        \begin{array}{ll}
        \underset{y}{\mbox{minimize}} & c(y), \\
        \mbox{subject to} & g(y; \theta) \leq 0, \\
        \end{array} \label{pb:optimization}
    \end{equation}
    where $c$ is a cost function, $g$ is a vector-valued inequality constraint, $y \in \mathbf{R}^{n_y}$ is the optimization variable, $\theta \in \mathbf{R}^{n_\theta}$ is the problem data. We do not include a dependency of the cost function $c$ on the problem data $\theta$. This would be an unnecessary complication as it is not required for the analysis of the collision detection Problem~\ref{pb:collision_detection} treated in this work.
    Differentiable optimization is interested in the sensitivities of the problem solution $y^*(\theta)$ and the optimal value of the problem $v^*(\theta) = c(y^*(\theta))$ with respect to the problem data $\theta$. 

    \textbf{Implicit Function Theorem} A general approach to obtain the sensitivities of $y^*$ with respect to $\theta$ is to apply the Implicit Function Theorem (IFT) to the Karush–Kuhn–Tucker (KKT) conditions~\cite{karush1939minima} of Problem~\ref{pb:optimization}. We denote $f$ the set of equality constraints contained in the KKT conditions. This is an implicit function of the form, 
    \begin{equation}
        f(y^*; \theta) = 0, \label{eq:implicit_function}
    \end{equation}
    where $f : \mathbf{R}^{n_y} \times \mathbf{R}^{n_\theta} \rightarrow \mathbf{R}^{n_y}$.

    At a solution point of \eqref{eq:implicit_function}, the sensitivities of the solution with respect to the problem data, i.e., $\partial y^* / \partial \theta$, can be computed utilizing the implicit function theorem under certain conditions\footnote{Namely, second-order sufficient conditions, strong complementary slackness, and linear independence of the active constraints~\cite{giorgi2018tutorial}.}~\cite{dini1907lezioni}.
    First, we expand Equation~\eqref{eq:implicit_function} to first order:
    \begin{equation}
        \frac{\partial f}{\partial y} \delta y + \frac{\partial f}{\partial \theta} \delta \theta = 0,
    \end{equation}
    and then solve for the relationship: 
    \begin{equation}
        \frac{\partial y^*}{\partial \theta} = -\Big(\frac{\partial f}{\partial y}\Big)^{-1} \frac{\partial f}{\partial \theta}. \label{eq:solution_sensitivity}
    \end{equation}
    This expression may still be useful even if $y$ is not exactly optimal ($y \approx y^*$), which is going to be the case in practice \cite{jaxopt_implicit_diff}.
    Often, solutions to \eqref{eq:implicit_function} are found using Newton's method and custom linear-system solvers can efficiently compute search directions for this purpose. Importantly, the factorization of $\partial f / \partial y$ used to find a solution during simulation can be reused to compute \eqref{eq:solution_sensitivity} at low computational cost requiring only back-substitution. Furthermore, each element of the problem-data sensitivity can be computed in parallel.

    \textbf{Sensitivity Analysis} If we are only interested in the sensitivity of the optimal value $v^*(\theta)$ with respect to $\theta$, we could use the IFT and apply the chain rule through the cost function:
    \begin{equation}
        \frac{d v^*}{d \theta} = \frac{\partial c }{\partial y} \frac{\partial y^*}{\partial \theta}.
        \label{eq:chain_rule}
    \end{equation}
    However, there is a simpler way to obtain this sensitivity by leveraging the optimal dual variables. In particular, under the same regularity conditions for the existence of $\partial y^\star/\partial \theta$, one can show that:
    \begin{equation}
        \frac{d v^*(\theta)}{d \theta} = \lambda^{*}(\theta)^T \frac{\partial g(y^*(\theta), \theta)}{\partial \theta},
    \label{eq:value_sensitivity}
    \end{equation}
    where $\lambda^*(\theta)$ are the optimal dual variables for problem data $\theta$ \cite{giorgi2018tutorial}.
 
    \subsection{Optimization-based Contact Simulation} \label{sec:contact_simulation}
        Inelastic contact between rigid objects can be simulated with a variational integration scheme~\cite{manchester2020variational}. It is expressed as a feasibility problem which results from the composition of the constrained \emph{Principle of Least Action} with the \emph{Maximum Dissipation Principle}~\cite{moreau2011unilateral}. For ease of exposition, we illustrate this feasibility problem for a rigid body making contact with its environment at a single point. Let $\mathcal{Q}$ denote the system's configuration space and $\Delta t >0$ the integrator time-step. Then, at each integration step, given the known current and previous configurations $q, q_{-} \in \mathcal{Q}$, we solve the following feasibility problem for the configuration at the next time step $q_{+}$:
        \begin{align}
            \mbox{find } & q_+, s_\gamma, s_\psi, s_\beta, \gamma, \psi, \beta 
            \label{pb:variational_integrator} \\
            \mbox{s. t. } & d(q_+; q, q_-, u) - N(p, n)^T \gamma - P(p, n)^T \beta = 0, 
            \label{eq:vi_1} \\
            & s_\gamma - \phi(q_+) = 0, 
            \label{eq:vi_2} \\
            & s_\psi - (\mu \gamma - \beta \ones) = 0, 
            \label{eq:vi_3} \\
            & s_\beta - \left( P(p, n) \left(\frac{q_+ - q}{\Delta t}\right) + \psi \ones \right) = 0, 
            \label{eq:vi_4} \\
            & s_\gamma \circ \gamma = 0, \quad s_\psi \circ \psi = 0, \quad s_\beta \circ \beta = 0, 
            \label{eq:vi_5} \\
            & s_\gamma, s_\psi, s_\beta, \gamma, \psi, \beta \succcurlyeq 0, \label{eq:vi_6}
        \end{align}
       
        where $\mu$ is the coefficient of friction, $u$ is the control input, $s_\gamma, s_\psi, s_\beta$ are slack variables, $\gamma, \psi, \beta$ are dual variables, $d$ is the manipulator equation, $\circ, \succcurlyeq$ are the generalized cone product and cone inequality respectively~\cite{vandenberghe2010cvxopt}. $P, N$ are contact Jacobians respectively mapping tangential and normal contact forces into generalized coordinates. These mappings only depend on the contact point $p$ and contact normal $n$, they do not require computing the Jacobian of the contact point location $\partial p / \partial q_{+}$.
        Equation~\eqref{eq:vi_1} encodes the system's dynamics, Equation~\eqref{eq:vi_2} the non-penetration constraint, Equation~\eqref{eq:vi_3} the friction cone constraint, Equation~\eqref{eq:vi_4}
        the maximum dissipation principle stating that friction forces oppose the movement of the contact point, Equations~\eqref{eq:vi_5},
        \eqref{eq:vi_6} are complementarity constraints.

        We note that contact simulation between a convex object and a convex environment requires three mappings, all taking as input the configuration $q \in \mathcal{Q}$ of the convex object:
        \begin{itemize}
        \item The Signed Distance Function (SDF) $\phi: \mathcal{Q} \rightarrow \mathbf{R}$ measures the distance between the object and the environment. Its sign indicates if the object is in contact or not with the environment: overlapping $\phi < 0$, in contact $\phi = 0$, separated $\phi > 0$.
        % \sumeet{For this ``preliminaries" section, this should be defined precisely as the distance map. This way your contact normal, defined as $\partial \phi/\partial x$ is of unit norm and orthgonal to the contact surface.} \addition{for the classical SDF, I am not sure how to get negative values when objects are overlapping}
        \item The contact point $p: \mathcal{Q} \rightarrow \mathbf{R}^3$, is the point at which the contact wrench is applied. It has to coincide with the actual point of contact when the object is in contact with the environment. However, its location is not restricted to the object's surface when the object is not in contact $\phi(q) \neq 0$ since no contact wrench is applied.
        \item The contact normal $n: \mathcal{Q} \rightarrow \mathbf{S}^2$, where $\mathbf{S}^2$ is the 2-sphere sub-manifold in $\mathbf{R}^3$, defines the direction along which impact forces are applied, friction forces being applied within the plane orthogonal to $n(q)$. 
        For convex objects, a valid definition for the contact normal is the direction orthogonal to a hyperplane separating the two objects. \addition{In this case, the contact normal belongs to the \textit{normal cones} of both objects \cite{leine2007stability}, \cite{brogliato1999nonsmooth}}. For a pair of objects in contact, the direction along which moving the object maximally increases the SDF is a valid contact normal. This is formally defined as $n(q) = \frac{\partial \phi}{\partial x}$ where $x$ is the position of the object. 
    \end{itemize}
        
    \subsection{Single-level Reformulations of Bilevel Problems} \label{sec:bilevel_problems}
    Bilevel optimization is a special type of optimization where a \emph{lower-level} problem is embedded within an \emph{upper-level} problem. 
    The bilevel solution method embeds the solution map of the lower-level problem into the KKT conditions of the upper-level problem.
    The drawbacks of this approach are that the solution map is costly to evaluate since it requires solving the lower-level problem and it is usually non-smooth (even under strong assumptions) \cite{beck2021gentle}. 

    The solution method used throughout this work relies on a single-level reformulation. It embeds the KKT conditions of the lower-level problem into the upper-level problem. 
    The single-level reformulation is equivalent to the original bilevel problem for convex lower-level problems that satisfy Slater’s constraint qualification \cite{beck2021gentle, dempe2012bilevel}. These assumptions are always respected in our collision detection context (see Problem~\ref{pb:collision_detection}).
    Finally, under the same assumptions one can embed the KKT conditions of the \emph{dual-form} of the lower-level problem instead \cite{beck2021gentle}. This is not a method that we have explored yet.

\section{Bilevel Optimization-based Simulation}
\label{sec:bilevel_optimization}

    In simple scenarios (e.g. plane-sphere), contact points and normals can be computed analytically. For more complex collision shapes, we resort to optimization-based collision detection.
    
    \subsection{Optimization-based Collision Detection}
    
\begin{figure}[t]
        \begin{center}
            \begin{tikzpicture}
                \draw (0, 0) node[inner sep=0] {\includegraphics[width=1.0\columnwidth]{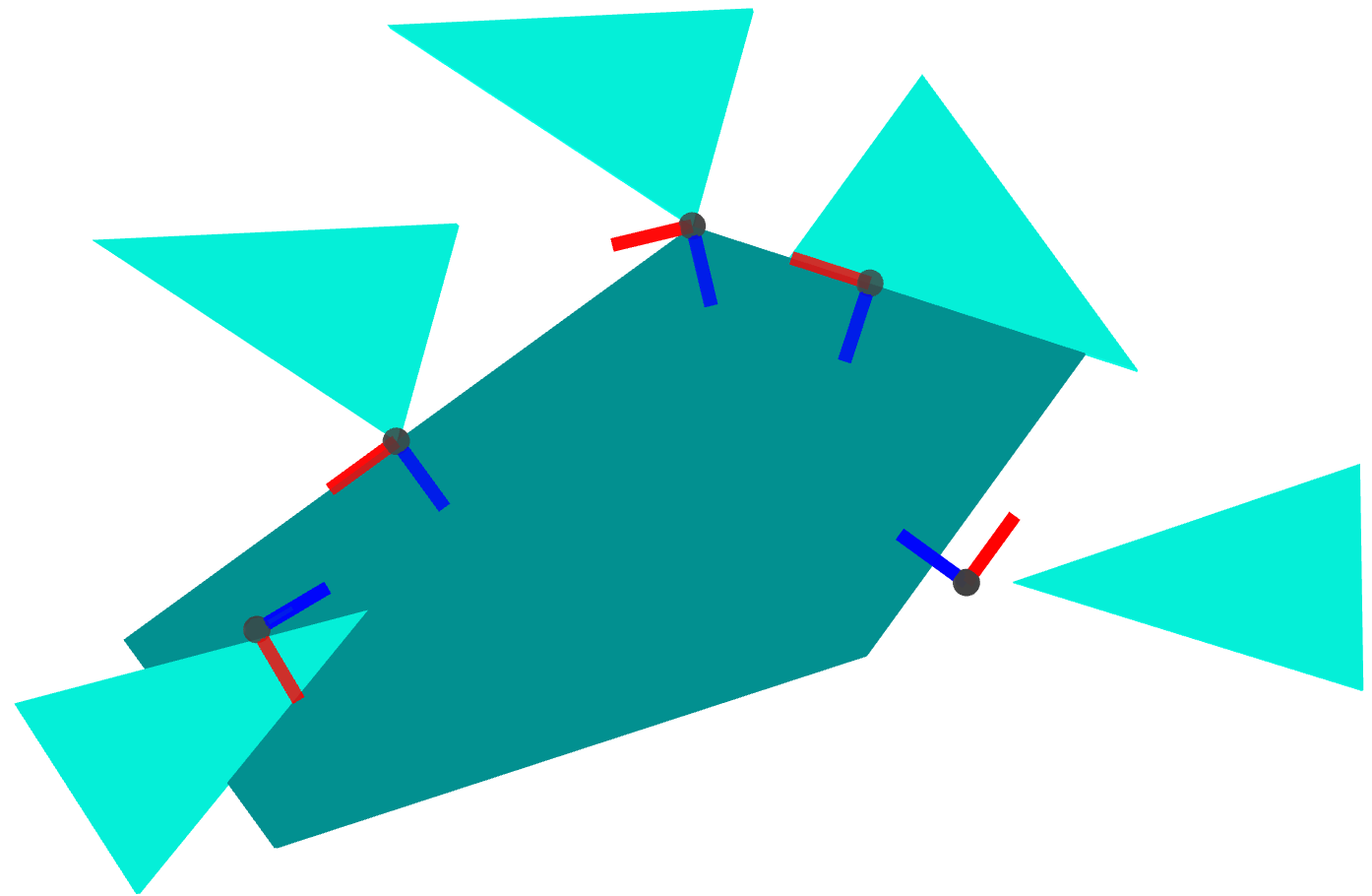}};
                \draw (-1.0*3.50, -1.0*1.90) node {$a$};
                \draw (-1.0*2.20, +1.0*0.90) node {$b$};
                \draw (-1.0*0.30, +1.0*2.30) node {$c$};
                \draw (+1.0*1.50, +1.0*1.50) node {$d$};
                \draw (+1.0*3.60, -1.0*0.80) node {$e$};
            \end{tikzpicture}
        \end{center}
        % \vspace{-3mm}
        \caption{We display the contact frames (contact point in gray, contact normal in blue, tangent in red) between objects with sharp edges. \addition{The collision detection Problem~\ref{pb:collision_detection} provides useful contact points and contact normals even when objects are not in contact ($\textbf{e}$). As highlighted in Section \ref{sec:contact_simulation}, 
        the contact point location only needs to be accurate when the objects are in contact.} Additionally, when objects are overlapping ($\textbf{a}$) the contact normal indicates a valid direction to split them apart. This formulation additionally returns meaningful contact points and contact normals when objects are in vertex-face contact ($\textbf{b}$), vertex-vertex contact ($\textbf{c}$), face-face contact ($\textbf{d}$).}
        \label{fig:contact_frames}
    \end{figure}
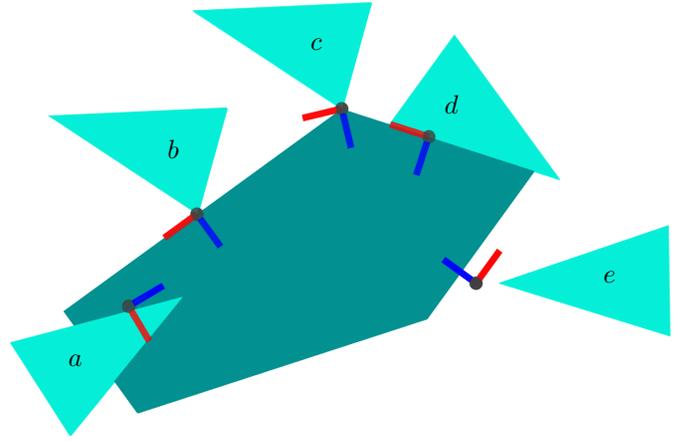

    The DCOL algorithm~\cite{tracy2022differentiable} formulates contact detection between convex shapes $\mathcal{O}_1$ and $\mathcal{O}_2$ as a convex program that respects Slater's constraint qualifications\footnote{Slater's constraint qualifications are verified for shapes that have non-zero relative interior in $\mathbf{R}^3$.}. Each shape $\mathcal{O}_i$ is encoded by a convex inequality constraint $\mathcal{O}_i = \{x \in \mathbf{R}^3 | g_i(x, \alpha, \epsilon_i; \theta) \leq 0\}$ where $\epsilon_i$ is an auxiliary variable required for shapes like cylinders or capsules, $\theta$ contains the shape parameters and its configuration in 3D space, $\alpha \in \mathbf{R_+}$ linearly scales the dimensions of the shape e.g. with $\alpha = 0$, $\mathcal{O}_i$ is scaled down to a single point in 3D space. 
    DCOL finds the minimum scaling $\alpha$ required to reduce the intersection between the two shapes to a single point $p$,
    \begin{equation}
        \begin{array}{ll}
        \underset{p, \alpha, \epsilon_1, \epsilon_2}{\mbox{minimize}} & \alpha, \\
        \mbox{subject to} & g_1(p, \alpha, \epsilon_1; \theta) \leq 0, \\
        & g_2(p, \alpha, \epsilon_2; \theta) \leq 0, \\
        & -\alpha \leq 0. \\
        \end{array} \label{pb:collision_detection}
    \end{equation}
    We can define the \addition{Pseudo Signed Distance Function\footnote{\addition{The ``distance'' we define here does not respect the triangle inequality.}} (P-SDF)} 
    % \sumeet{Don't call this the SDF, since it isn't.} 
    between $\mathcal{O}_1$ and $\mathcal{O}_2$ as $\phi = \alpha - 1$. It is easily verified that $\phi$ is negative when shapes are overlapping, and positive otherwise. This measure is closely related to the notion of growth distances \cite{ong1996growth}, \cite{ahmadigeometry}. Figure~\ref{fig:contact_frames} illustrates the collision detection results obtained with this approach. The gradient $\partial \phi/\partial x$ defines a valid contact normal (proof in Appendix~\ref{app:contact_normal}), yet $\partial \phi/\partial x$ may not be unit norm, therefore re-scaling is required to obtain a unit-norm contact normal $n \propto \partial \phi/\partial x$. 

    % \sumeet{Show $p$ when in contact is same as SDF $p$. Show $\partial \phi/\partial x$ is orthogonal to one of the contact surfaces, and therefore is a valid contact normal. Add remark: Note that $\partial \phi/\partial x$ may not be unit norm (since it is not the actual SDF), therefore re-scaling is required to appropriately define the friction Jacobian $P$, and forces $\beta$.}
    
    \subsection{The Bilevel Simulation}
    Bilevel contact simulation combines the contact dynamics Problem~\ref{pb:variational_integrator} (upper-level) and the collision detection Problem~\ref{pb:collision_detection} (lower-level). To solve the bilevel problem, we evaluate and differentiate the KKT conditions of the upper-level problem. Each evaluation of these KKT conditions requires to solve the lower-level problem to obtain collision information i.e. the P-SDF $\phi$, contact point location $p$ and contact normal $n$. 
    This bilevel solution method, introduced in Section~\ref{sec:bilevel_problems}, is used in recent works \cite{tracy2022differentiable, wang2022linear}.
    
\section{Single-level Contact Simulation and Collision Detection}
\label{sec:silico}

    In this section, we present our novel formulation of contact dynamics unifying contact simulation and collision detection in a single-level optimization problem. We start-off by proposing a simpler way to differentiate through collision detection problems. Then, we illustrate the shortcomings of the bilevel formulation~\cite{tracy2022differentiable} on a simple scenario. Finally, we introduce our single-level formulation addressing the weaknesses of the bilevel approach.
    
\subsection{Contact Detection Sensitivity Analysis}
    For clarity, we detail the case of a single convex rigid body making contact with a convex-shaped environment (e.g. a plane). However, the results presented here naturally extend to multiple non-convex rigid bodies making contact with a non-convex environment assuming each non-convex shape is decomposed as a set of convex shape primitives as illustrated in Section~\ref{sec:experiments}.
    
    The solution map of Problem~\ref{pb:collision_detection} parameterized by the object's configuration ($\theta = q$) provides the contact point mapping $p(q)$ and the P-SDF mapping $\phi(q) = \alpha(q) - 1$. However, the contact normal $n$ requires differentiation of the solution $\alpha$ with respect to the object's position $x$; $n(q) \propto {\frac{\partial \phi}{\partial x}}^T = {\frac{\partial \alpha}{\partial x}}^T$. Previously, that was obtained via the IFT~\cite{tracy2022differentiable, tracy2022diffpills, wang2022linear}. However, a simpler method further exploits the structure of Problem~\ref{pb:collision_detection}. We remark that $\alpha$ is both a variable and the value we optimize for. Using the latter interpretation, we can leverage the optimal value sensitivity result given in Equation~\eqref{eq:value_sensitivity},
    \begin{equation}
        n \propto \left(\frac{\partial \alpha^*}{\partial x}\right)^T = \left(\frac{\partial g}{\partial x}\right)^T \lambda,
        \label{eq:normal_sensitivity}
    \end{equation} 
    where $\lambda$ is the dual variable associated with the inequality constraint $g$ defined as,
    \begin{equation}
        g(p, \alpha, \epsilon_1, \epsilon_2; q) = 
        \begin{bmatrix}
            g_1(p, \alpha, \epsilon_1; q) \\
            g_2(p, \alpha, \epsilon_2; q) \\
            -\alpha 
        \end{bmatrix} \in \mathbf{R}^{n_g},
        \label{eq:g_function}
    \end{equation}  
    where $q$ is the configuration of the system. 
    Compared to the IFT-based computation of the contact normal, our method is  significantly less computationally intensive. Our approach requires a simple matrix-vector product and evaluates the contact normal in linear time $O(n_g)$ (Eq.~\eqref{eq:normal_sensitivity}). In comparison, the IFT (Eq.~\eqref{eq:solution_sensitivity}) has a quadratic computational complexity $O(n_g^2)$ assuming the matrix has been pre-factorized, and cubic complexity $O(n_g^3)$ otherwise. In Section~\ref{sec:experiments}, we verify the computational benefits of our approach on a practical contact simulation scenario. 

   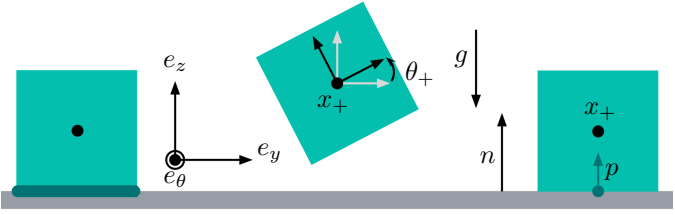
\begin{figure}[t]
        \begin{center}
            \begin{tikzpicture}
                \draw (0, 0) node[inner sep=0] {\includegraphics[width=1.0\columnwidth]{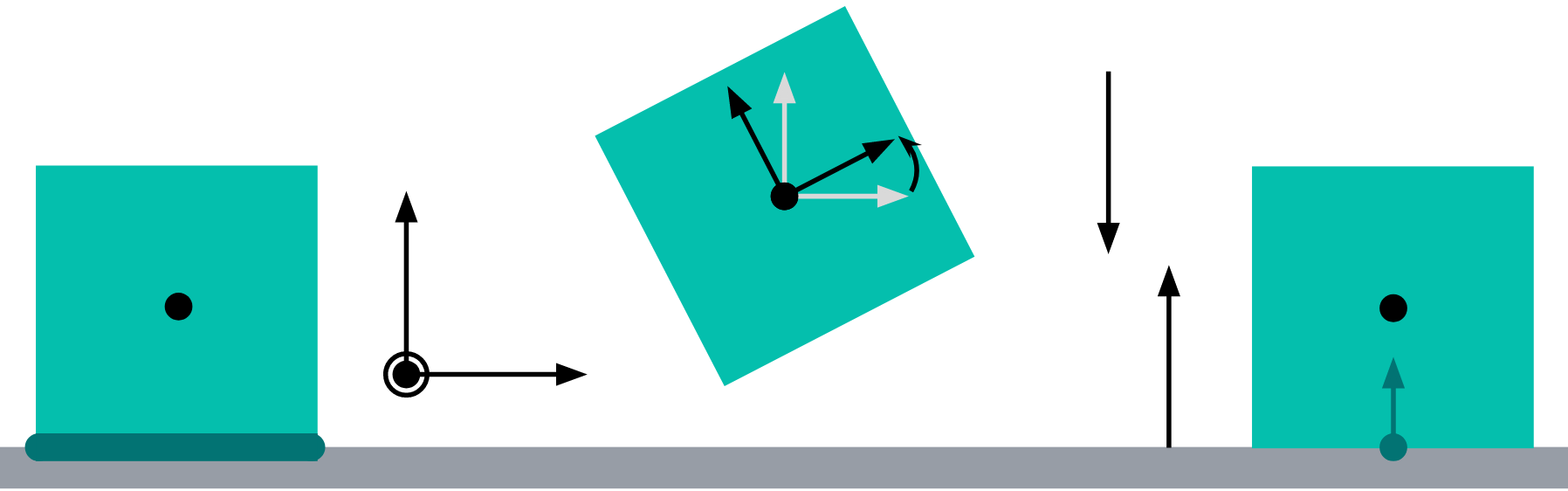}};
                \draw (+1.65, +0.60) node {$g$};
                \draw (+2.00, -0.70) node {$n$};
                \draw (-0.90, -0.65) node {$e_y$};
                \draw (-2.15, +0.55) node {$e_z$};
                \draw (-2.15, -1.00) node {$e_\theta$};
                \draw (+1.10, +0.40) node {$\theta_+$};
                \draw (-0.05, -0.00) node {$x_+$};
                \draw (+3.65, -0.90) node {$p$};
                \draw (+3.50, -0.10) node {$x_+$};
                % \draw (+0.60, +0.80) node {$g$};
                % \draw (+1.55, -0.60) node {$n$};
                % \draw (-0.15, -0.65) node {$e_y$};
                % \draw (-1.50, +0.80) node {$e_z$};
                % \draw (+3.40, -0.90) node {$p$};
                % \draw (+3.50, -0.10) node {$x_+$};
            \end{tikzpicture}
		\end{center}
        \caption{Left, in contact simulation scenarios, it is often the case that two shape primitives are in face-to-face contact. In such case, the collision detection problem has a continuum of solutions. Any point in the dark turquoise area is a valid contact point. \addition{Middle, the box pose is parameterized by a position $x_+ \in \mathbb{R}^2$ and orientation $\theta_+ \in \mathbb{R}$.} Right, solving for the contact point $p$ and the contact force jointly resolves this ambiguity. Indeed, there is a single contact point lying below the center of mass $x_+$ of the blue box.}
        \label{fig:contact_ambiguity}
    \end{figure}

    \subsection{Contact Point Non-Uniqueness}
    Optimization-based collision detection is a powerful method. However, when collision detection is considered in isolation as with the bilevel formulation, it exhibits non-uniqueness issues. This issue occurs for any face-to-face contact. We illustrate this with a simple 2D scenario: a box sitting on flat ground (Fig.~\ref{fig:contact_ambiguity}). The collision detection problem is expressed as follows, 

    % \begin{equation}
    %     \begin{array}{ll}
    %     \underset{p, \alpha}{\mbox{minimize}} & \alpha, \\
    %     \mbox{subject to} & A(p - x_{+}) - \alpha b \leq 0, \\
    %     & p^T n \leq 0, \\
    %     & -\alpha \leq 0. \\
    %     \end{array} \label{pb:box_collision_detection}
    % \end{equation}
    \addition{
    \begin{equation}
        \begin{array}{ll}
        \underset{p, \alpha}{\mbox{minimize}} & \alpha, \\
        \mbox{subject to} & AR(\theta_+)(p - x_{+}) - \alpha b \leq 0, \\
        & p^T n \leq 0, \\
        & -\alpha \leq 0. \\
        \end{array} \label{pb:box_collision_detection}
    \end{equation}
    where $p$ is the contact point, $A = [-e_z, e_y, e_z, -e_y]^T \in \mathbf{R}^{4 \times 2}$ and $b = \ones$ define the box shape, $n = e_z$ is the normal to the ground, $x_{+} = [0, 1]$ and $\theta_+ = 0$ are the position and orientation of the box in 2 dimensions, $R(\theta_+)$ rotates vectors from the world frame to the box frame.}
    The KKT conditions of Problem~\ref{pb:box_collision_detection} are as follows: 
    \begin{align}
        r_{\mbox{col}}(v_{\mbox{col}}) = \begin{bmatrix}
            R^T A^T \gamma_p + \gamma_n n \\
            1 - \gamma_p^T b - \gamma_\alpha \\
            s_p + AR (p - x_{+}) - b \alpha \\
            s_n + p^T n\\
            s_\alpha - \alpha\\
            \gamma_p \circ s_p\\
            \gamma_n \circ s_n\\
            \gamma_\alpha \circ s_\alpha
        \end{bmatrix} = 0, 
    \end{align}
    where $v_{\mbox{col}} = [p, \alpha, \gamma_p, \gamma_n, \gamma_\alpha, s_p, s_n, s_\alpha]$. The Jacobian of the KKT conditions, 
    \begin{align}
        \frac{\partial r_{\mbox{col}}}{\partial v_{\mbox{col}}} = \begin{bmatrix}
        0 & 0 & R^TA^T & n & 0 & 0 & 0 & 0 \\
        0 & 0 & -b^T & 0 & -1 & 0 & 0 & 0 \\
        AR & -b & 0 & 0 & 0 & I_4 & 0 & 0 \\
        n^T & 0 & 0 & 0 & 0 & 0 & 1 & 0\\
        0 & -1 & 0 & 0 & 0 & 0 & 0 & 1\\
        0 & 0 & s_p & 0 & 0 & \gamma_p & 0 & 0\\
        0 & 0 & 0 & s_n & 0 & 0 & \gamma_n & 0\\
        0 & 0 & 0 & 0 & s_\alpha & 0 & 0 & \gamma_\alpha\\
        \end{bmatrix}.
    \end{align}   
    When the box is lying on the ground, this Jacobian is rank-deficient\footnote{
    At a solution point,
    $p = [0, 0], \alpha = 1, \gamma_p = [1,0,0,0], \gamma_n = 1, \gamma_\alpha = 0, s_p= [0, 1, 2, 1], s_n = 0, s_\alpha = 1$, the $1^{\mbox{st}}$, $11^{\mbox{th}}$ and $13^{\mbox{th}}$ columns are linearly dependent.
    } illustrating the fact that any point $p$ located on the boundary between the box and the floor is an acceptable collision point (Fig.~\ref{fig:contact_ambiguity} left).
    
    More generally, when two shape primitives are in face-to-face contact, there is a continuum of valid contact points for the collision detection problem. Isolating the collision detection problem artificially introduces ambiguity in the solution. This is problematic since typically only a single contact point is dynamically-valid within the continuum of collision detection solutions. 
    % \sumeet{Add sentence on consequences of this - if you have non-uniquenes, you can just pick one location (according to some simple heuristic) and send it to the simulation problem. If it is not being differentiated, does it matter?}
    Considering collision detection in conjunction with the contact dynamics problem resolves this ambiguity. We establish this formally in the simple box-on-ground scenario in Section~\ref{sec:single_level_formulation}, we further verify this empirically for a set of complex scenarios in Section~\ref{sec:experiments}.
    In the box-on-ground scenario, there is a unique contact point that intersects the box and the ground while maintaining a stable contact. This is the point located below the center of mass (Fig.~\ref{fig:contact_ambiguity} right). When considering the collision problem in isolation, we lack information to identify the right point among all valid contact points.

\subsection{Single-Level Formulation}
    \label{sec:single_level_formulation}

    Equipped with a simple expression for the contact normal (Eq.~\eqref{eq:normal_sensitivity}), we can formulate a single-level optimization problem. It combines the KKT conditions of the contact dynamics Problem~\ref{pb:variational_integrator} with the KKT conditions of the collision detection Problem~\ref{pb:collision_detection}\footnote{
    One could imagine obtaining the contact normal by applying the IFT (Eq.~\eqref{eq:chain_rule}) to the collision detection problem and embedding this complex expression in the single-level KKT conditions. However, this would require the computation of prohibitively expensive third-order tensors to evaluate the Jacobian of the KKT system. 
    }. We obtain a \emph{composed optimization problem} solving for the contact forces and contact location jointly; 
    \begin{equation}
        \begin{array}{ll}
        \mbox{find} & q_+, s_\gamma, s_\psi, s_\beta, \gamma, \psi, \beta,\\
            &\phi, p, n, \alpha, \epsilon_1, \epsilon_2, s, \lambda \\
        \mbox{subject to} & d(q_+; q, q_-, u) - N(p, n)^T \gamma - P(p, n)^T \beta = 0, \\
        & s_\gamma - \phi(q_+) = 0, \\
        & s_\psi - (\mu \gamma - \beta \ones) = 0, \\
        & s_\beta - \left( P(p, n) \left(\frac{q_+ - q}{\Delta t}\right) + \psi \ones \right) = 0, \\
        & s_\gamma \circ \gamma = 0, \quad s_\psi \circ \psi = 0, \quad s_\beta \circ \beta = 0, \\
        & s_\gamma, s_\psi, s_\beta, \gamma, \psi, \beta \succcurlyeq 0,\\
        & \phi = \alpha - 1, \\
        & n = \lambda^T \nabla_{x} g / || \lambda^T \nabla_{x} g||_2, \\
        & 1 + \lambda^T \nabla_\alpha g = 0, \\
        & \lambda^T \nabla_p g = 0, \\
        & \lambda^T \nabla_{\epsilon_1} g = 0, \\
        & \lambda^T \nabla_{\epsilon_2} g = 0, \\
        & s + g = 0, \\
        & s \circ \lambda = 0, \\
        & s, \lambda \succcurlyeq 0,\\
        \end{array} \label{pb:single_level_problem}
    \end{equation}
    where $s$ is the slack variable associated with the inequality constraint $g(p, \alpha, \epsilon_1, \epsilon_2; q_{+}) \leq 0$. Note that the collision detection constraint $g$ is evaluated at the next time step configuration $q_+$. 
    We solve this nonlinear complementarity problem with a primal-dual interior point method \cite{mehrotra1992implementation}.
    With a bilevel formulation, Problem~\ref{pb:variational_integrator} is solved using the solution map of Problem~\ref{pb:collision_detection} to compute the P-SDF $\phi(q)$ and the contact point $p(q)$, and the solution map is differentiated to get the contact normal $n(q)$. \addition{We observe that both formulations involve nonlinear collision detection and that a solution to the single-level formulation is also a solution to the bilevel formulation.}

    Yet, compared to a bilevel optimization formulation, our approach has several advantages. 
    
    \begin{figure}[t]
		\begin{center}
            \begin{tikzpicture}
                \draw (0, 0) node[inner sep=0] {\includegraphics[width=1.0\columnwidth]{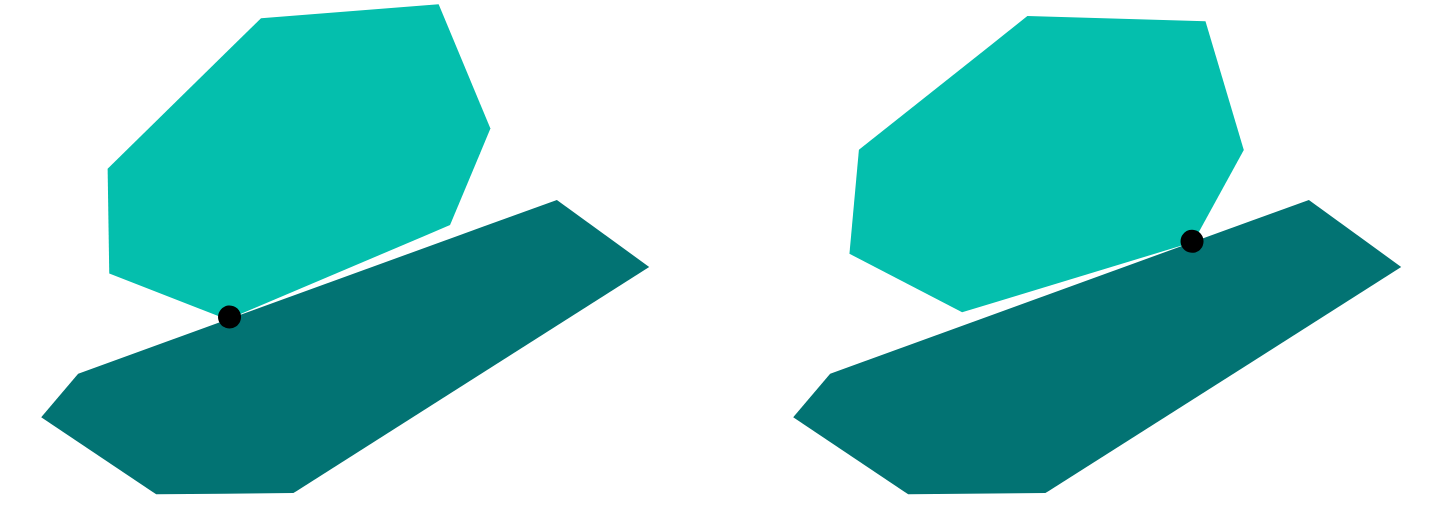}};
                \draw (+3.80, +0.50) node {$p(q+\delta q)$};
                \draw (-3.75, -0.35) node {$p(q)$};
                \draw (-2.60, +0.60) node {$\mathcal{O}_1$};
                \draw (-2.60, -0.70) node {$\mathcal{O}_2$};
                \draw (+2.00, +0.60) node {$\mathcal{O}_1$};
                \draw (+2.00, -0.70) node {$\mathcal{O}_2$};
            \end{tikzpicture}
        \end{center}
		\caption{The contact point between two shape primitives $\mathcal{O}_1$ and $\mathcal{O}_2$ can change significantly for arbitrarily small relative pose changes, i.e. the contact location $p(q)$ is a discontinuous function of the configuration $q$.}
		\label{fig:contact_position_discontinuity}
	\end{figure}
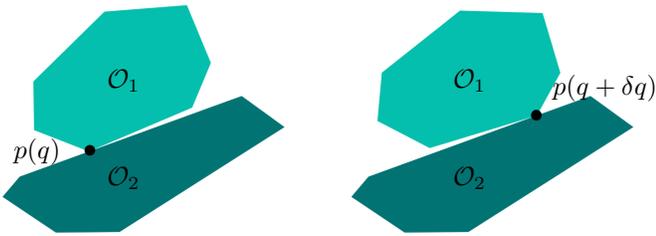
    
    In terms of computational stability, the reliance of the bilevel approach on the solution map of the collision detection is not desirable. Indeed, while the collision detection Problem~\ref{pb:collision_detection} is convex, its solution map can be a discontinuous function of the configuration of the system $q$. This happens when two shape primitives are in face-to-face contact (Fig.~\ref{fig:contact_position_discontinuity}). This is rare in the trajectory optimization setting~\cite{zhang2020optimization}, and contact simulation under zero gravity~\cite{tracy2022differentiable}. However, it is ubiquitous in most practical contact simulation scenarios (i.e. non-zero gravity). 
    As a consequence, the bilevel formulation attempts to find the root of a set of discontinuous equations. On the contrary, the single-level formulation relies on continuous expressions to compute the contact data $\phi, p, n$ (Pb.~\ref{pb:single_level_problem}), thus avoiding discontinuous KKT conditions \cite{beck2021gentle}. \addition{This is why applying an interior point method \cite{mehrotra1992implementation} i.e. performing root-finding on a sequence of relaxed problems is much more reliable with the single-level formulation.}
    
    On the computational complexity front, our method does not require to solve the collision detection problem multiple times like the bilevel formulation. We solve it once jointly with the contact dynamics problem. We assess computational benefits in a simple contact simulation scenario (Section~\ref{sec:experiments}). 

    Finally, we concretely illustrate the benefits of this formulation with the box-on-ground scenario. 
    The Jacobian of the constraints of the single-level Problem~\ref{pb:single_level_problem} simply concatenates the Jacobian of the KKT conditions of the collision detection problem with a few more rows and columns corresponding to additional contact dynamics equations and variables respectively,  
    \begin{equation}
        \frac{\partial r}{\partial v} =
        \begin{bmatrix} 
            \begin{matrix}
                \quad \cdot \: \\
                \quad \cdot \: \\
                \quad \cdot \: \\
                \quad \cdot \: \\
            \end{matrix} & \vline & 
            \begin{matrix}
                0 & & 0 \\
                -\gamma e_y^T & & 0 \\
                0 & & 0 \\
                0 & & -1 \\
            \end{matrix} & \vline & 
            \begin{matrix}
                \cdot \\
                \cdot \\
                \cdot \\
                \cdot \\
            \end{matrix} \\
            \hline
            \begin{matrix}
                \quad \cdot \: \\
                \quad \cdot \: \\
                \quad \cdot \: \\
            \end{matrix} & \vline & 
            \begin{matrix}
                \quad & \quad \\
                \quad & \frac{\partial r_{\mbox{col}}}{\partial v_{\mbox{col}}} \\
                \quad & \quad \\
            \end{matrix}
        \end{bmatrix},
        \label{eq:simplified_jacobian}
    \end{equation}   
    where $r$ and $v$ are the constraints and variables of the single-level formulation, $\cdot$ denotes omitted entries, $\gamma$ is the impact force applied by the ground on the box. 
    The $-\gamma e_y^T$ entry is the Jacobian of the angular momentum conservation equation with respect to the contact point location $p$. Thanks to this entry, the Jacobian $\frac{\partial r}{\partial v}$ is full rank. Thus, factoring in dynamics equations can resolve rank deficiencies of the collision detection problem.
    The intuition behind this is that, only the contact point located below the center of mass is \emph{physically valid} and maintains the box at rest on the floor. Other \emph{geometrically valid} contact points would generate a torque resulting in a rotation of the box. 
    Further details about the box-on-ground single-level formulation are provided in Appendix~\ref{sec:box_on_ground}. 

    % \footnote{
    % The matrix remains full-rank as long as the solution requires a non-zero contact force $\gamma$ (e.g. non-zero gravity scenario). Otherwise, this is not an issue since the contact has no effect on the movement of the box, so the contact point location does not matter. I think this footnote will be confusing for the reader if anything. 
    % } 

    \subsection{Simulation Differentiability}
    We can differentiate the contact simulation optimization problem with respect to the problem data. Similar to the original formulation used in Dojo \cite{howelllecleach2022}, we use the IFT to obtain the dynamics gradients. For instance, we can obtain the sensitivity of the next configuration $q_+$ of a robot with respect to problem data such as its initial conditions, its mass, inertia and the parameters of its collision shapes. 

    The exact contact dynamics are not smooth and applying the IFT to an exact solution of Problem~\ref{pb:single_level_problem} would provide subgradients of the dynamics. These gradients are not capturing the local dynamics landscape very well making them of little use for optimization through contact problems \cite{pang2022global}. 
    Similarly to Dojo \cite{howelllecleach2022}, we relax the complementarity constraints $a \circ b = 0$ in Problem~\ref{pb:single_level_problem} with a relaxation parameter $\rho \in \mathbf{R}_+$,
    \begin{align}
        a \circ b = \rho \mathbf{e}
        \label{eq:relaxation_parameter}
    \end{align}
    where $\circ$ denotes the cone product, $\mathbf{e}$ is the neutral element for the cone product \cite{vandenberghe2010cvxopt}. 
    Applying IFT with relaxed complementarity constraints generates smoother gradients that better approximate the local dynamics landscape. These relaxed gradients have demonstrated their effectiveness in many optimization-through-contact problems: reinforcement learning, model predictive control, global planning, system identification and trajectory optimization \cite{howelllecleach2022, lecleach2021fast, pang2022global}.
    
    Introducing complementarity relaxation in the contact dynamics smooths impacts. For collision detection, complementarity relaxation effectively \emph{soften corners and contact normals} of sharp shapes like polytopes and cones. This allows for smooth dynamics gradient computation even when simulating interaction between sharp shapes with discontinuous contact normals.

    Importantly, our approach allows for accurate contact dynamics simulation with infinitesimal complementarity relaxation levels $\rho = 10^{-6}$. Additionally, it can provide gradients at any desired level of smoothness by differentiating the solution obtained at a user-specified relaxation level (typically $\rho \in [10^{-6}, 10^{-2}]$).    
    \begin{figure}[t]
        \begin{center}
            \includegraphics[width=1.0\columnwidth]{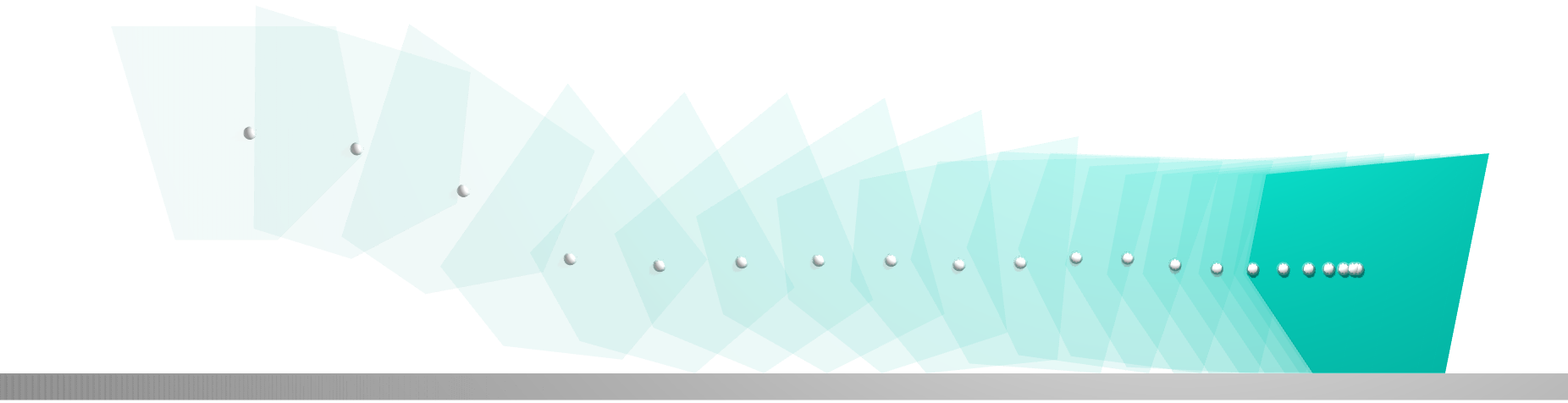}
        \end{center}
        \caption{We drop a polytope-shaped object on flat ground, this simple benchmark scenario highlights the face-to-face contact issue. We quantitatively evaluate the amount of ground penetration and the simulation robustness (Fig.~\ref{fig:performance_matrix}).}
        \label{fig:polytope_drop}
    \end{figure}

    \begin{figure}[t]
        \begin{center}
            \begin{tikzpicture}
                \draw (0, 0) node[inner sep=0] {\includegraphics[width=1.0\columnwidth]{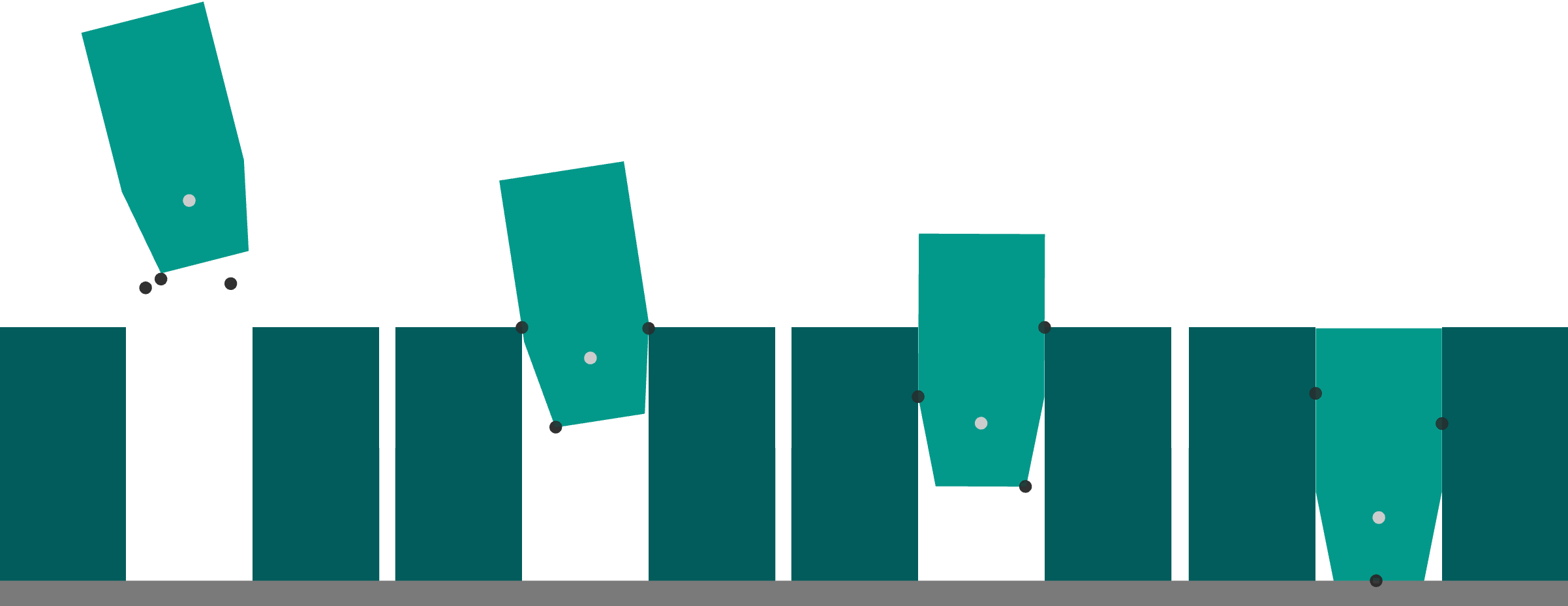}};
                \draw (-3.35, -1.9) node {$t = 0.0$};
                \draw (-1.10, -1.9) node {$t = 0.5$};
                \draw (+1.10, -1.9) node {$t = 1.0$};
                \draw (+3.35, -1.9) node {$t = 2.0$};
            \end{tikzpicture}
        \end{center}
        \caption{Peg-in-hole \cite{newman2001interpretation} insertion task, the peg makes contact with the sides and the floor at three points (black dots).}
        \label{fig:peg_in_hole}
    \end{figure}
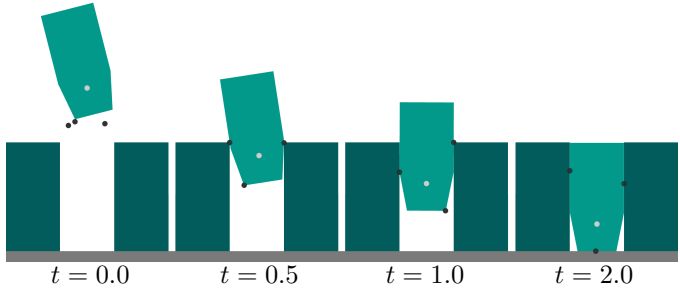
    
    \begin{figure}[t]
        \begin{center}
            \includegraphics[width=0.7\columnwidth]{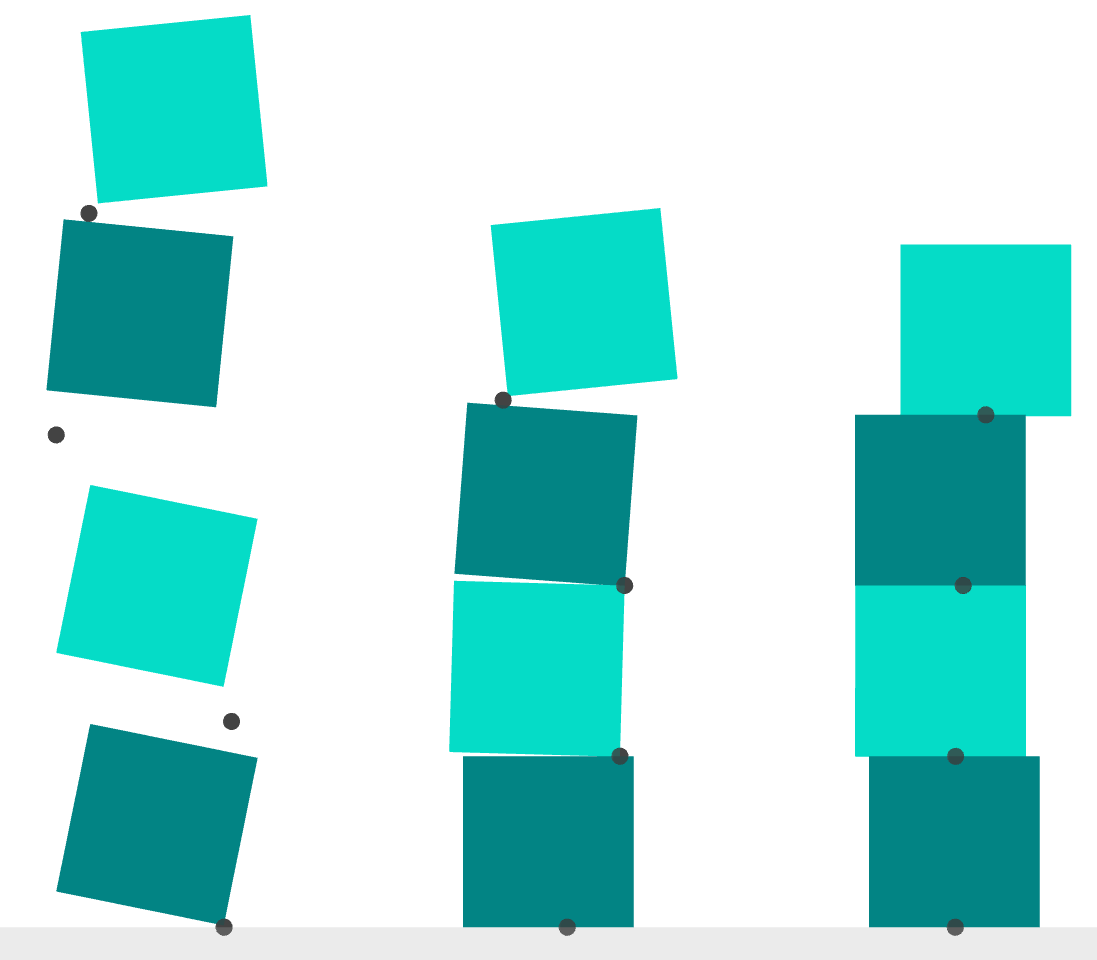}
            \hfill
            \includegraphics[width=0.23\columnwidth]{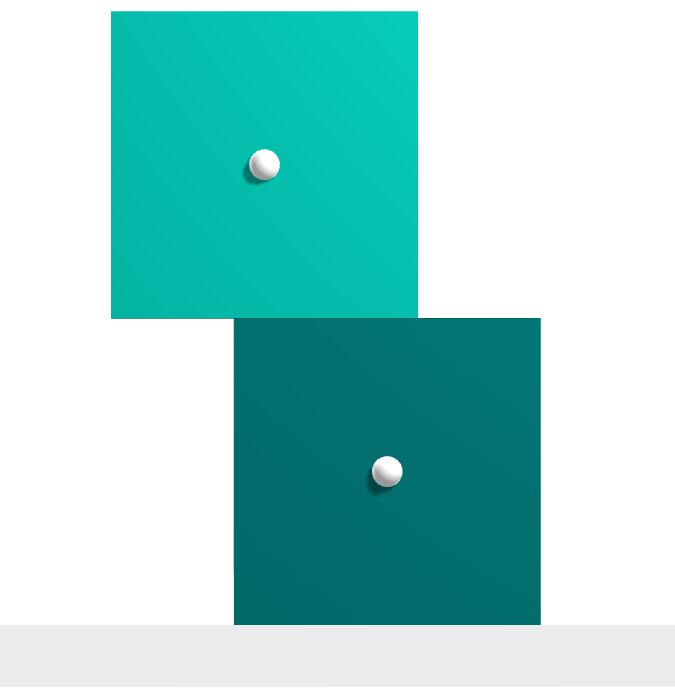}
        \end{center}
        \caption{Left, we simulate a stack of 4 blocks, our single-level formulation identifies physically meaningful contact points (black dots) allowing for stable stack simulation.
        Right, the block stacking scenario; we initialize the two blocks in a stable stack configuration. We simulate the system forward in time and verify quantitatively that the stack is stable (Fig.~\ref{fig:momentum_matrix}).}
        \label{fig:stacked_blocks}
    \end{figure}

\section{Experiments}
\label{sec:experiments}
    We highlight the capabilities of the single-level formulation on a set of simulation examples. Then, we quantitatively evaluate the benefits of the single-level approach in comparison to a bilevel formulation, in terms of simulation accuracy and reliability. Finally, we leverage the differentiability of the simulator to solve a series of optimization-through-contact problems. 

    \subsection{Simulation Scenarios}
    \textbf{Convex bundles} First, we illustrate how we can compose convex shape primitives (polytopes, cylinders, planes, capsules, and ellipsoids) to build complex non-convex shapes (Appendix \ref{app:shape_primitives}). Shape primitives can be easily combined though a union operation forming potentially non-convex shapes dubbed as ``convex bundles'' (Fig.~\ref{fig:convex_bundles}).
    To simulate contact between this convex bundle and the ground, we compute contact points between each convex primitive and the ground. 
    
    We can also use the Minkowski sum of several convex primitives to form a complex convex shape. We demonstrate this with two primitives $\mathcal{O}_1$ and $\mathcal{O}_2$, 
    \begin{align}
        g_1(p, \alpha, \epsilon_1; \theta) \leq 0, \\
        g_2(p, \alpha, \epsilon_2; \theta) \leq 0,
    \end{align}
    reusing notation from Problem~\ref{pb:collision_detection}.
    The inequality constraint encoding the Minkowski sum is then, 
    \begin{align}
        g_{12}(p, \alpha, \epsilon; \theta) =
        \begin{bmatrix}
            g_1(p_1, \alpha, \epsilon_1; \theta)\\
            g_2(p - p_1, \alpha, \epsilon_2; \theta)\\
        \end{bmatrix}
        \leq 0,
    \end{align}
    where $\epsilon = [\epsilon_1^T, \epsilon_2^T, p_1]^T$. The generated shape is convex and can be used as a primitive to generate even more complex shapes through unions and Minkowski sums (Fig.~\ref{fig:convex_bundles}).     
    
    \textbf{Face-to-Face Contacts}
    One challenging simulation situation occurs when two objects are in face-to-face contact. These events occur very often in robotics scenarios. For instance: dropping an object onto flat ground (Fig.~\ref{fig:polytope_drop}), stacking blocks to form a tower (Fig.~\ref{fig:stacked_blocks}), performing peg-in-hole \cite{newman2001interpretation} insertion (Fig.~\ref{fig:peg_in_hole}).

    \begin{figure}[t]
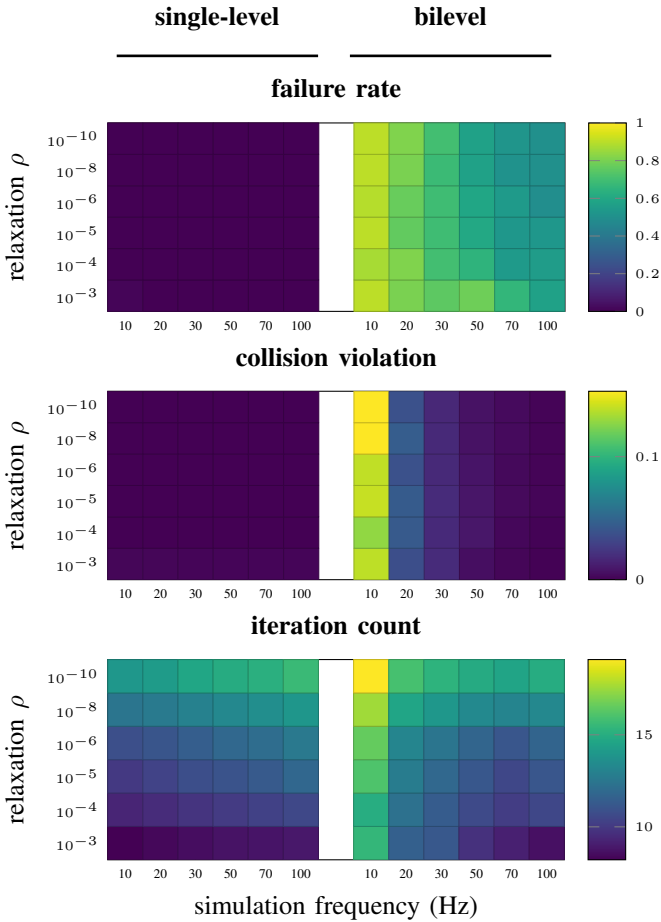

        \begin{raggedleft}
            \begin{minipage}{.30\columnwidth}
                \centering
                \textbf{single-level}
                \rule{\columnwidth}{1pt}
            \end{minipage}
            \hspace{0.3cm}%
            \begin{minipage}{.30\columnwidth}
                \centering
                \textbf{bilevel}
                \rule{\columnwidth}{1pt}
            \end{minipage}%
            \hspace{1.55cm}
            \vskip0.5em
            
            \includegraphics[width=1.0\columnwidth, height=0.40\columnwidth]{tikz/failure_rate.tikz}
            \includegraphics[width=1.0\columnwidth, height=0.40\columnwidth]{tikz/violation.tikz}
            \includegraphics[width=0.99\columnwidth, height=0.48\columnwidth]{tikz/iteration_count.tikz}
        \end{raggedleft}
        \caption{
        We run 3600 object-dropping simulations (Fig.~\ref{fig:polytope_drop}) with a series of complementarity relaxation values (smaller means stiffer contact), and a series of simulation frequencies. 
        For each simulation, we count the number of iterations taken by the NCP solver, we measure the maximum collision violation between the object and the floor, we assess if the solve was successful i.e. each simulation step required less than 30 iterations to complete. Our single-level solver experiences 4 failures across the 3600 runs. The bilevel approach fails between $55\%$ and $95\%$ of the time and generates large collision violations especially at low simulation frequencies. For both approaches, smaller relaxation parameter values (i.e. stiffer contact) increase the iteration count. 
        }
        \label{fig:performance_matrix}
    \end{figure}
   
    \subsection{Baseline Comparison}
    We quantitatively compare the single-level formulation to the bilevel approach. 

    \textbf{Simulation Robustness}
    First, we assess the simulation reliability on a simple object dropping scenario (Fig.~\ref{fig:polytope_drop}). This scenario tests the ability of the simulator to handle face-to-face contact reliably. 
    We randomly sample initial conditions, we simulate the system for $1.5$s. Because of gravity, the convex object will eventually make a face-to-face contact with the floor (Fig.~\ref{fig:polytope_drop}). We compare the bilevel and our single-level approach over a wide range of simulation frequencies, and values for relaxation $\rho$, we observe (i) significantly lower failure rate ($50\%$ down to $0.1\%$), (ii) less collision violation (more than a $100 \times$ smaller on average), and (iii) particularly lower iteration count at low simulation frequencies of 10 Hz (Fig.~\ref{fig:performance_matrix}).

    \begin{figure}[t]
        \begin{raggedleft}
            \begin{minipage}{.30\columnwidth}
                \centering
                \textbf{single-level}
                \rule{\columnwidth}{1pt}
            \end{minipage}
            \hspace{0.3cm}%
            \begin{minipage}{.30\columnwidth}
                \centering
                \textbf{bilevel}
                \rule{\columnwidth}{1pt}
            \end{minipage}%
            \hspace{1.55cm}
            \vskip0.5em
            \includegraphics[width=1.0\columnwidth, height=0.48\columnwidth]{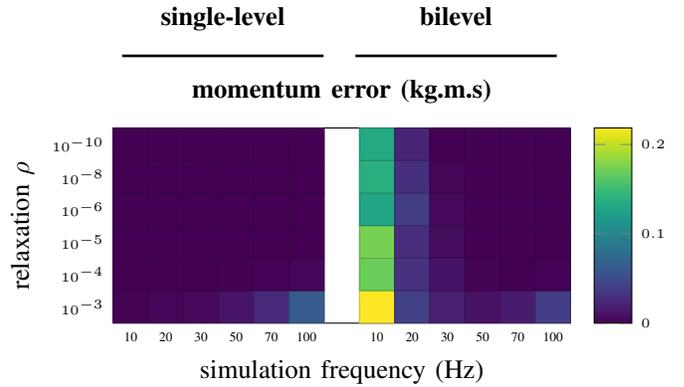}
        \end{raggedleft}
        \caption{We stack two blocks on top of each other in 10 different ways forming a stable tower (Fig.~\ref{fig:stacked_blocks} right). We simulate the system for $1.0$s using the bilevel and our single-level approach. To evaluate the ability to simulate stacked blocks accurately, we measure the average momentum drift during the simulation. As the tower is initially stable, the momentum is supposed to remain null. We observe that the bilevel simulation is often very inaccurate (blocks jittering, tower collapsing) especially at low simulation frequencies i.e. simulation large time steps. On the contrary, the single-level simulation is stable for all 360 runs, with a momentum error an order of magnitude smaller on average.
        }
        \label{fig:momentum_matrix}
    \end{figure}
            
    \textbf{Momentum Conservation}
    Then, we test the simulation accuracy on a simple object stacking scenario (Fig.~\ref{fig:stacked_blocks}). Because of the face-to-face contact, the bilevel formulation struggles to simulate the system stably. It generates spurious movements for the blocks, often resulting in the tower collapsing. The single-level formulation successfully simulates this scenario. We quantitatively evaluate the simulation accuracy by measuring the error in terms of momentum (Fig.~\ref{fig:momentum_matrix}). The single-level formulation significantly outperforms the bilevel approach both visually and quantitatively ($10 \times$ smaller momentum error).

    \begin{figure}[t]
        \begin{center}
            \includegraphics[width=1.0\columnwidth, height=0.40\columnwidth]{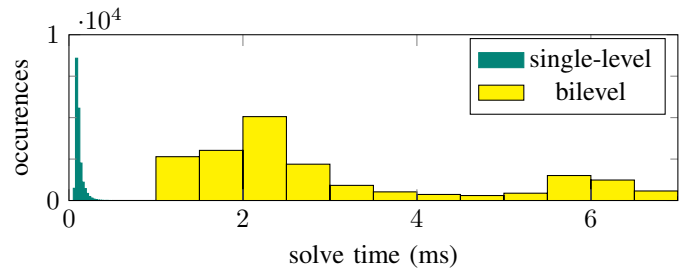}
        \end{center}
        \caption{We simulate a block falling on the floor (Fig.~\ref{fig:polytope_drop}) with the single-level and the bilevel formulations. We record the solve time for $20\;000$ simulation steps. The mean solve time is on average $30$ times larger for the bilevel formulation compared to the single-level formulation.}
        \label{fig:solve_times}
    \end{figure}

    \textbf{Computational Complexity} 
    We simulate the box falling on a plane with the single-level and bilevel methods. We compare the simulation times for both approaches. The single-level approach is $30$ times faster on average (Fig.~\ref{fig:solve_times}). We attribute these computational gains to several factors. 
    For a single simulation step, the bilevel method requires to solve the collision detection problem multiple times. The single-level formulation solves it only once jointly with the contact dynamics problem. 
    Applying Newton's method to the bilevel formulation requires costly computation of the contact point and the signed distance function gradients with the IFT. The single-level formulation does not require such gradient information.

    \subsection{Optimization-Through-Contact}
    We illustrate the benefits of the differentiable simulation framework on a challenging optimization-through-contact problem. We focus on a contact-rich manipulation task requiring to manipulate an object to a desired location with two spherical fingertips (Fig.~\ref{fig:rrt}). We solve this problem thanks to a RRT-based method \cite{pang2022global}. This method extensively leverages smooth gradient information to compute robot control inputs and to define a distance metric. We found that the method worked best with a significant amount of smoothing $\rho = 3\cdot 10^{-3}$. With our proposed approach the relaxation parameter $\rho$ (Eq.~\eqref{eq:relaxation_parameter}) provides smooth gradient through the contact dynamics (e.g. impact, friction) and the collision detection (e.g. contact point location, contact normal). This is appealing since quantities such as contact normals are often discontinuous functions of the robot configuration. Our formulation can provide smooth gradients through these discontinuities. We solve the grasping task in $1$ second on a laptop, the plan generated showcases a non-trivial contact sequence akin to the complex plans demonstrated by Pang et al. \cite{pang2022global}. 
    
\section{Limitations}
\label{sec:limitations}

Our method does have some limitations. 
    (i) The interior-point method used to solve Problem~\ref{pb:single_level_problem} is difficult to warm-start. It is a well-known limitation of interior-point methods \cite{yildirim2002warm, john2008implementation}. In the simulation context, reusing the solution obtained at the previous time step to warm-start the solver can deliver huge computational gains. We have implemented a simple warm-starting strategy reusing the previous solution with a small perturbation on the slack and dual variables $x \leftarrow x + 10^{-2} \mathbf{e}$. While this strategy significantly cuts down solve times, it slightly decreases the reliability of the solver. Further investigation would be necessary to provide a more robust warm-starting strategy. 
    (ii) As the method yields a single contact point between a pair of convex objects, torsional friction is not modeled. Including torsional friction would require adding a scalar inequality constraint to the contact dynamics Problem~\ref{pb:single_level_problem}.

\section{Conclusion}
\label{sec:conclusion}

    In this work, we have presented a new formulation unifying contact dynamics and collision detection in a single-level optimization problem. This approach yields drastically improved simulation robustness and accuracy compared to previous methods. We demonstrate our method on a series of robotics simulation scenarios. Additionally, the differentiability of the contact formulation allows it to be efficiently integrated into policy and trajectory optimization frameworks such as RL and MPC, respectively.  
    Future work will explore broad-phase collision detection \cite{mirtich1996impulse} detecting and pruning out inactive collision pairs to further speed-up simulation times. Additionally, extending the Julia \cite{bezanson2017julia} implementation to PyTorch \cite{pytorch} or JAX \cite{bradbury2018jax} would enable a highly parallelized contact simulation with GPU support.

\section*{Acknowledgments}
    The authors would like to thank Kevin Tracy and Taylor Howell for useful and insightful technical discussions on collision detection; Tetiana Parshakova for her helpful advice on sensitivity analysis; \addition{Michael Posa for his deeply constructive feedback on an earlier version of this work.}

\bibliographystyle{ieeetr}
\bibliography{main}

\clearpage
\appendix

\subsection{Contact Normal via Sensitivity Analysis}
    \label{app:contact_normal}
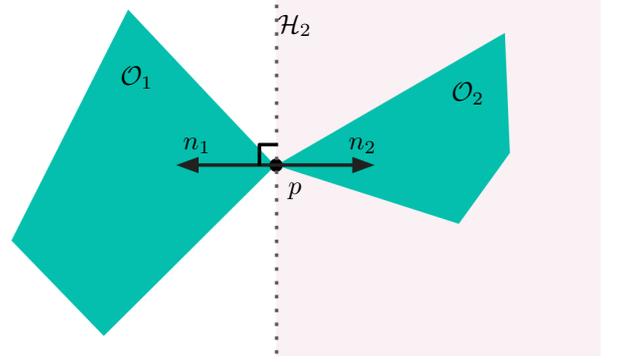
\begin{figure}[t]
        \begin{center}
            \begin{tikzpicture}
                \draw (0, 0) node[inner sep=0] {\includegraphics[width=1.0\columnwidth]{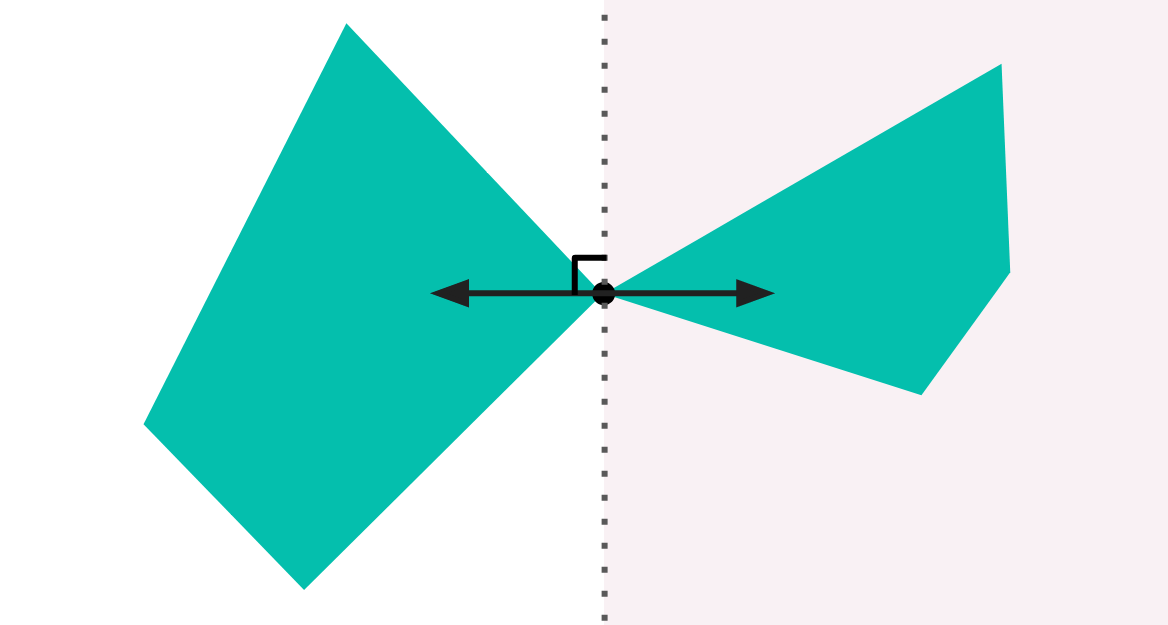}};
                \draw (+0.40, -0.20) node {$p$};
                \draw (-0.90, +0.40) node {$n_1$};
                \draw (+1.30, +0.40) node {$n_2$};
                \draw (-1.70, +1.30) node {$\mathcal{O}_1$};
                \draw (+2.70, +1.10) node {$\mathcal{O}_2$};
                \draw (+0.40, +2.00) node {$\mathcal{H}_2$};
                % \draw (-1.90, -0.70) node {$x_1$};
                % \draw (+2.40, -0.00) node {$x_2$};
            \end{tikzpicture}
        \end{center}
        \caption{When two convex shapes $\mathcal{O}_1$ and $\mathcal{O}_2$ are in contact, the contact normal $n_1 = - n_2$ provided by sensitivity analysis defines a separating hyperplane (dotted line). 
        }
        \label{fig:separating_hyperplane}
    \end{figure}

    Intuitively, a contact normal between two convex objects should be orthogonal to the surface of both objects at the contact point. However, when contact surfaces are not smooth (e.g. polytope) orthogonality to the surface is ambiguous. To accommodate for this technical difficulty, we define a valid contact normal as a direction orthogonal to a hyperplane separating the two convex objects (Fig.~\ref{fig:separating_hyperplane}). 
    
    \textbf{Proof:} We demonstrate that sensitivity analysis provides a valid contact normal when the objects are in contact i.e. $\alpha^* = 1$.
    The KKT conditions of Problem~\ref{pb:collision_detection}:
    \begin{align}
        1  - \lambda^*_\alpha + {\lambda_1^*}^T \nabla_\alpha g_1 + {\lambda_2^*}^T \nabla_\alpha g_2 &= 0 \\
        {\lambda_1^*}^T \nabla_p g_1 + {\lambda_2^*}^T \nabla_p g_2 &= 0\\
        \lambda_1^* \nabla_{\epsilon_1} g_1 &= 0 \\
        \lambda_2^* \nabla_{\epsilon_2} g_2 &= 0 \\
        s_\alpha^* - \alpha &= 0 \\
        s_1^* + g_1 &= 0 \\
        s_2^* + g_2 &= 0 \\
        \lambda_\alpha^* \circ s_\alpha^* &= 0 \\
        \lambda_1^* \circ s_1^* &= 0 \\
        \lambda_2^* \circ s_2^* &= 0 \\
        \lambda_\alpha^*, \lambda_1^*, \lambda_2^*, s_\alpha^*, s_1^*, s_2^* &\succcurlyeq 0
    \end{align}
    with object $\mathcal{O}_i$ described by $g_i(p, \alpha, \epsilon_i; x_i) \leq 0$ where $x_i$ is the 3D position of object $i$. Since the position object $\mathcal{O}_i$ factors solely into inequality $g_i$ as $g_i(p, \alpha, \epsilon_i; x_i) = f_i(p-x_i, \alpha, \epsilon_i)$, we have:
    \begin{align}
        {\lambda^*}^T \nabla_{x_i} g = {\lambda^*_i}^T \nabla_{x_i} g_i = -{\lambda^*_i}^T \nabla_{p} g_i, \quad \forall i \in \{1,2\},
    \end{align}
    where $g$ is defined in Equation~\eqref{eq:g_function} and $\lambda^* = [{\lambda^*_1}^T {\lambda^*_2}^T {\lambda^*_\alpha}^T]^T$.
    
    To complete the proof, we show that the hyperplane orthogonal to $n \propto {\nabla_p g_2}^T \lambda_2^*$ is a separating hyperplane. We start off by introducing the half-space $\mathcal{H}_2$ (Fig.~\ref{fig:separating_hyperplane}) defined as, \begin{align}
        g_{\mathcal{H}_2}(p, \alpha) &= A_{\mathcal{H}_2} (p - p^*) + b_{\mathcal{H}_2} (1 - \alpha) \\
        A_{\mathcal{H}_2} &= {\lambda_2^*}^T \nabla_p g_2 \\
        b_{\mathcal{H}_2} &= - {\nabla_\alpha g_2}^T \lambda_2^*.
    \end{align}
    If we swap object $\mathcal{O}_2$ with half-space $\mathcal{H}_2$; we can verify that the KKT conditions of Problem~\ref{pb:collision_detection} are still verified with the same $\alpha^* = 1$ and contact point $p^*$\footnote{For completeness, the optimal dual and slack variables are $\lambda_{\mathcal{H}_2} = 1$, $s_{\mathcal{H}_2} = 0$.}.
    Thus, the P-SDF between $\mathcal{O}_1$ and $\mathcal{H}_2$ is $\phi = \alpha^* -1 = 0$. This means that $\mathcal{O}_1$ and $\mathcal{H}_2$ are in contact but do not overlap\footnote{Formally, the intersection of $\mathcal{O}_1$ and $\mathcal{H}_2$ has no relative interior in $\mathbf{R}^3$.}. 
    By symmetry, a similar argument can be made about $\mathcal{O}_2$ and the half-space $\mathcal{H}_1$ orthogonal to $n$. Thus, the hyperplane orthogonal to $n$ and intersecting $p^*$ is indeed a separating hyperplane.

    \textbf{Practical considerations:}
    Following the proof, we can provide two equivalent ways of computing the contact normal.
    \begin{align}
        n \propto - {\nabla_{x_1} g_1}^T \lambda^*_1 = {\nabla_{x_2} g_2}^T \lambda^*_2.
        \label{eq:equivalent_normal}
    \end{align}
    These two quantities are equal at a solution point, however during the solve they are not necessarily equal. In practice, using $n_1 \propto {\nabla_{x_1} g_1}^T \lambda_1$ to map contact wrenches into object $1$'s generalized coordinates and $n_2 \propto {\nabla_{x_2} g_2}^T \lambda_2$ for object $2$, significantly improves the solver convergence properties. 

    In Problem~\ref{pb:collision_detection}, we solve for a \emph{single} contact point $p$. This point is located on the surface of both objects when they are in contact, inside both objects when they are overlapping, outside when they are separated. Equal and opposite contact wrenches are applied on both objects at this single contact point. It would be physically meaningless to apply contact wrenches at a contact point located outside both objects. Fortunately, at solution points of Problem~\ref{pb:single_level_problem} objects are not overlapping and wrenches are non-zero only when the contact point is located on the surface of both objects.

\subsection{Shape Primitives and Contact Normals} \label{app:shape_primitives}
    We provide the contact normal expressions for a set of shape primitives.
        
    \begin{itemize}
        \item Plane:
    \end{itemize}
    \begin{align}
        g(p, \alpha; x) &= a^T (p - x) - b \leq 0\\
        n &\propto - \lambda^* a
    \end{align}
    
    \begin{itemize}
        \item Polytope:
    \end{itemize}
    \begin{align}
        g(p, \alpha; x) &= A(p - x) - \alpha b \leq 0\\
        n &\propto - {\lambda^*}^T A
    \end{align}
    
    \begin{itemize}
        \item Padded Polytope:
    \end{itemize}
    \begin{align}
        g(p, \alpha, \epsilon; x) &= \begin{bmatrix}
            A(p + \epsilon - x) - \alpha b \\
            \epsilon^T \epsilon - \alpha^2 R^2
        \end{bmatrix} \leq 0 \\
            n &\propto {\lambda^*}^T \begin{bmatrix}
            - A \\
            0_{1 \times 3}
        \end{bmatrix}
    \end{align}
    
    \begin{itemize}
        \item Ellipsoid:
    \end{itemize}
    \begin{align}
        g(p, \alpha; x) &= (p - x)^T E^T E (p-x) - \alpha^2 \leq 0\\
        n &\propto 2{\lambda^*}^T E^T E (x-p^*)
    \end{align}
    
    \begin{itemize}
        \item Capsule:
    \end{itemize}
    \begin{align}
        g(p, \alpha, \epsilon; x) &= \begin{bmatrix}
            ||p - (x + \epsilon v)||_2 - \alpha R \\
            \epsilon - \alpha L/2\\
            -\epsilon - \alpha L/2
        \end{bmatrix} \leq 0 \\
            n &\propto {\lambda^*}^T \begin{bmatrix}
            \frac{-(p - (x + \epsilon v))}{||p - (x + \epsilon v)||_2} \\
            0_{2 \times 3} \\
        \end{bmatrix}
    \end{align}
    
    \begin{itemize}
        \item Cylinder:
    \end{itemize}
    \begin{align}
        g(p, \alpha, \epsilon; x) &= \begin{bmatrix}
            ||p - (x + \epsilon v)||_2 - \alpha R \\
            \epsilon - \alpha L/2\\
            -\epsilon - \alpha L/2\\
            -p + (x - \alpha \frac{L}{2} v)^T v \\
            p - (x + \alpha \frac{L}{2} v)^T v
        \end{bmatrix} \leq 0 \\
            n &\propto {\lambda^*}^T \begin{bmatrix}
            \frac{-(p - (x + \epsilon v))}{||p - (x + \epsilon v)||_2} \\
            0_{2 \times 3} \\
            v^T\\
            -v^T
        \end{bmatrix}
    \end{align}
    
    \begin{itemize}
        \item Cone:
    \end{itemize}
    \begin{align}
        \hat{p} &= p - x + \alpha \frac{3H}{4} v \\
        g(p, \alpha; x) &= \begin{bmatrix}
            ||\hat{p}_{2:3}||_2 - \mbox{tan}(\nu) \hat{p}_1 \\
            (p - x - \alpha \frac{H}{4} v)^T v
        \end{bmatrix} \leq 0 \\
            n &\propto {\lambda^*}^T \begin{bmatrix}
            - \mbox{tan}(\nu) x_1 &\frac{\hat{p}_{2:3}^T}{||\hat{p}_{2:3}||_2} \mbox{Diag}(x_{2:3}) \\
            v_1 & v_{2:3}^T
        \end{bmatrix}
    \end{align}

\subsection{Detailed Box-On-Ground Scenario} \label{sec:box_on_ground}
    In the box-on-ground scenario, the collision detection problem is expressed by Problem \ref{pb:box_collision_detection} with Lagrangian defined as follows, 
    % as follows, 
    % \begin{equation}
    %     \begin{array}{ll}
    %     \underset{p, \alpha}{\mbox{minimize}} & \alpha, \\
    %     \mbox{subject to} & AR(\theta_+)(p - x_{+}) - \alpha b \leq 0, \\
    %     & p^T n \leq 0, \\
    %     & -\alpha \leq 0. \\
    %     \end{array} \label{pb:full_box_collision_detection}
    % \end{equation}
    % where $x_+$ and $\theta_+$ are the position and orientation of the box in 2 dimensions, $R(\theta_+)$ rotates vectors from the world frame to the box frame. 
    % The Lagrangian associated with Problem~\ref{pb:box_collision_detection},
    \begin{align}
        L = &\alpha 
        + \gamma_p^T \left[s_p + AR (p - x_{+}) - \alpha b \right] \notag\\
        &+ \gamma_n^T \left[s_n + p^T n \right]
        + \gamma_\alpha^T \left[ s_\alpha - \alpha \right].
    \end{align}
    The KKT conditions of Problem~\ref{pb:box_collision_detection}, concatenated with the equality constraints of the contact dynamics (Pb.~\ref{pb:variational_integrator}), 
    \begin{align}
        r(v) = \begin{bmatrix}
            \frac{m}{\Delta t} (x_+ -2x + x_-) - \gamma n - m g \Delta t\\
            \frac{J}{\Delta t} (\theta_+ -2\theta + \theta_-) - \gamma e_y (p - x_+)\\
            s_\gamma - (\alpha - 1)\\ 
            \gamma \circ s_\gamma\\
            R^T A^T \gamma_p + \gamma_n n \\
            1 - \gamma_p^T b - \gamma_\alpha \\
            s_p + AR (p - x_{+}) - b \alpha \\
            s_n + p^T n\\
            s_\alpha - \alpha\\
            \gamma_p \circ s_p\\
            \gamma_n \circ s_n\\
            \gamma_\alpha \circ s_\alpha
        \end{bmatrix} = 0, 
    \end{align}
    where $v = [x_+, \theta_+, \gamma, s_\gamma, p, \alpha, \gamma_p, \gamma_n, \gamma_\alpha, s_p, s_n, s_\alpha]$, $m$ is the mass, $J$ is the inertia, $\Delta t$ is the time step, $g$ is the gravity field, $e_y = [1, 0]$ is the horizontal unit vector, $\gamma$ is the impact force applied by the ground on the box, $s_\gamma$ is the slack variable associated with the impact constraint, the $+$ and $-$ subscripts indicate next and previous time steps respectively. The dynamics equation are discretized with a first-order variational integrator. 

    With the initial conditions setting the box a rest on the ground $x = x_- = [0, 1]$, $\theta = \theta_- = 0$.
    The solution to this set of constraints is $x_+ = [0, 1]$, $\theta_+ = 0$, $p = [0, 0], \alpha = 1, \gamma_p = [1,0,0,0], \gamma_n = 1, \gamma_\alpha = 0, s_p= [0, 1, 2, 1], s_n = 0, s_\alpha = 1$

    A simplified version of the single-level constraints Jacobian is presented in Equation~\eqref{eq:simplified_jacobian}. For completeness, we provide the detailed expression,
    \begin{align}
        & \quad \frac{\partial r}{\partial v} =  \\
        & \begin{bsmallmatrix}
            \frac{m}{\Delta t} I_2 & \cdot & n & \cdot & \cdot & \cdot & \cdot & \cdot & \cdot & \cdot & \cdot & \cdot \\
            \frac{J}{\Delta t} & \gamma e_y^T & e_y^T(p - x_+) & \cdot & -\gamma e_y^T & \cdot & \cdot & \cdot & \cdot & \cdot & \cdot & \cdot \\
            \cdot & \cdot & \cdot & 1 & \cdot & -1 & \cdot & \cdot & \cdot & \cdot & \cdot & \cdot \\
            \cdot & \cdot & s_\gamma & \gamma & \cdot & \cdot & \cdot & \cdot & \cdot & \cdot & \cdot & \cdot \\
            \cdot & \frac{\partial R^T A^T \gamma_p^T}{\partial \theta_+} & \cdot & \cdot & \cdot & \cdot & R^TA^T & n & \cdot & \cdot & \cdot & \cdot \\
            -AR & \frac{\partial AR(p - x_+)}{\partial \theta_+} & \cdot & \cdot & \cdot & \cdot & -b^T & \cdot & -1 & \cdot & \cdot & \cdot \\
            \cdot & \cdot & \cdot & \cdot & AR & -b & \cdot & \cdot & \cdot & I_4 & \cdot & \cdot \\
            \cdot & \cdot & \cdot & \cdot & n^T & \cdot & \cdot & \cdot & \cdot & \cdot & 1 & \cdot\\
            \cdot & \cdot & \cdot & \cdot & \cdot & -1 & \cdot & \cdot & \cdot & \cdot & \cdot & 1\\
            \cdot & \cdot & \cdot & \cdot & \cdot & \cdot & s_p & \cdot & \cdot & \gamma_p & \cdot & \cdot\\
            \cdot & \cdot & \cdot & \cdot & \cdot & \cdot & \cdot & s_n & \cdot & \cdot & \gamma_n & \cdot\\
            \cdot & \cdot & \cdot & \cdot & \cdot & \cdot & \cdot & \cdot & s_\alpha & \cdot & \cdot & \gamma_\alpha\\
        \end{bsmallmatrix} \notag
    \end{align}   
    Contrary to the collision detection Jacobian, this physics-informed Jacobian is full-rank at the solution point. This precludes the existence of a continuum of solutions around the solution point.

\end{document}